\documentclass[mathpazo]{cicp}
\usepackage[numbers]{natbib}
\usepackage{subfig}
\usepackage{tikz}
\usepackage[linesnumbered,ruled,vlined]{algorithm2e}
\usepackage{booktabs} 
\usepackage[colorlinks,urlcolor=blue,linkcolor=blue,citecolor=blue]{hyperref}
\usepackage{lineno}
\usepackage{amssymb,amsmath}
\let\oldcitet=\citet
\let\oldcitep=\citep 
\renewcommand{\citet}[1]{\textcolor[rgb]{0,0,1}{\oldcitet{#1}}}
\renewcommand{\citep}[1]{\textcolor[rgb]{0,0,1}{\oldcitep{#1}}}
\usepackage{lineno}

\begin{document}
\title{Investigating and Mitigating Failure Modes in Physics-informed Neural Networks (PINNs)}

\author[S.~Basir]{Shamsulhaq Basir\corrauth}
\address{Mechanical Engineering and Materials Science Department at the University of Pittsburgh, Pittsburgh, PA 15261, USA.}
\email{{\tt shb105@pitt.edu,shamsbasir@gmail.com} (S.~Basir)}

\begin{abstract}
This paper explores the difficulties in solving partial differential equations (PDEs) using physics-informed neural networks (PINNs). PINNs use physics as a regularization term in the objective function. However, a drawback of this approach is the requirement for manual hyperparameter tuning, making it impractical in the absence of validation data or prior knowledge of the solution. Our investigations of the loss landscapes and backpropagated gradients in the presence of physics reveal that existing methods produce non-convex loss landscapes that are hard to navigate. Our findings demonstrate that high-order PDEs contaminate backpropagated gradients and hinder convergence. To address these challenges, we introduce a novel method that bypasses the calculation of high-order derivative operators and mitigates the contamination of backpropagated gradients. Consequently, we reduce the dimension of the search space and make learning PDEs with non-smooth solutions feasible. Our method also provides a mechanism to focus on complex regions of the domain. Besides, we present a dual unconstrained formulation based on Lagrange multiplier method to enforce equality constraints on the model's prediction, with adaptive and independent learning rates inspired by adaptive subgradient methods. We apply our approach to solve various linear and non-linear PDEs.
\end{abstract}


\maketitle
\section{Introduction}
\label{sec:introduction}
A wide range of physical phenomena can be explained with partial differential equations (PDEs), including sound propagation, heat and mass transfer, fluid flow, and elasticity. The most common methods (i.e., finite difference, finite volume, finite element, spectral element) for solving problems involving PDEs rely on domain discretization. Thus, the quality of the mesh heavily influences the solution error. Moreover, mesh generation can be tedious and time-consuming for complex geometries or problems with moving boundaries. While these numerical methods are efficient for solving forward problems, they are not well-suited for solving inverse problems, particularly data-driven modeling. In this regard, neural networks can be viewed as an alternative meshless approach to solving PDEs. 

\citet{dissanayake1994neural} introduced neural networks as an alternative approach to solving PDEs. The authors formulated a composite objective function that aggregated the residuals of the governing PDE with its boundary condition to train a neural network model. Independent from work presented in \cite{dissanayake1994neural}, \citet{van1995neural} also proposed a similar approach for the solution of a two-dimensional magnetohydrodynamic plasma equilibrium problem. Several other researchers adopted the work in \cite{dissanayake1994neural,van1995neural} for the solution of nonlinear Schrodinger equation \cite{monterola2001solving}, Burgers equation \cite{hayati2007feedforward}, self-gravitating N body problems \cite{quito2001solving}, and chemical reactor problem \cite{Parisi2003solving}. Unlike earlier works, \citet{lagaris1998artificial} proposed to create trial functions for the solution of PDEs that satisfied the boundary conditions by construction. However, their approach is not suitable for problems with complex geometries. It is possible to create many trial functions for a particular problem. But to choose an optimal trial function is a challenging task, particularly for PDEs. 

Recently, the idea of formulating a composite objective function to train a neural network model following the approach in \cite{dissanayake1994neural,van1995neural} has found a resurgent interest thanks to the works in\cite{weinan2017proposal,Raissi2019,sirignano2018dgm}. This particular way of learning the solution to strong forms of PDEs is commonly referred to as physics-informed neural networks (PINNs) \cite{Raissi2019}. PINNs employ physics as a regularizing term in their objective function. However, this approach brings forth the challenge of manually adjusting the corresponding hyperparameters. Furthermore, the absence of validation data or prior knowledge of the solution to the Partial Differential Equation (PDE) can render PINNs impracticable. The deep Ritz method has been proposed to solve variational problems arising from PDEs \cite{weinan2017proposal}. This method enforces boundary conditions through a hyperparameter that cannot be tuned without validation data or prior knowledge of the solution. Thus, it is not well-suited for solving forward problems.There is a growing interest in using neural networks to learn the solution to PDEs \cite{raissi2019deep,kissas2020machine,mao2020physics,gao2021phygeonet,PATEL2022110754,PECANN_2022,RANS_PINN2022,jagtap2021extended}. Despite the great promise of PINNs for the solution of PDEs, several technical issues remained a challenge, which we discuss further in section \ref{sec:PINNs}. 
Different from the earlier approaches \cite{dissanayake1994neural,van1995neural,Parisi2003solving,Raissi2019}, we recently proposed \emph{physics and equality constrained artificial neural networks} (PECANNs) that are based on constrained optimization techniques. Furthermore, we used a maximum likelihood estimation approach to seamlessly integrate noisy measurement data and physics while strictly satisfying the boundary conditions. In section \ref{sec:Proposed_Method}, we discuss our proposed formulation, constrained optimization problem, and unconstrained dual problem.

Our contribution is summarized as follows:
\begin{itemize}
    \item We demonstrate and investigate failure modes in physics-informed neural networks for the solution of partial differential equations by comparing its learning process to a purely data-driven baseline model.
    \item We conduct a sensitivity analysis of backpropagated gradients in the presence of physics and show that high-order PDEs contaminate the backpropagate gradients by amplifying the inherent noise in the predictions of a neural network model, particularly during the early stages of training.
    \item We propose to precondition high-order PDEs by introducing auxiliary variables. We then learn the solution to the primary and auxiliary variables using a single neural network model to promote learning shared hidden representations and mitigate the risk of overfitting.
    
    \item We propose an unconstrained dual formulation by adapting the Lagrange multiplier method following the ideas from adaptive subgradient methods. Our formulation is meshless, geometry invariant, and tightly enforces boundary conditions and conservation laws without manual tuning. 
    \item We solve several benchmark problems and demonstrate several orders of magnitude improvements over existing physics-informed neural network approaches.
\end{itemize}

\section{Technical Background}
\label{sec:tech_background}
This section aims to highlight and discuss the technical ingredients of physics-based neural networks. Let us start by considering a partial differential equation of the form
\begin{align}
    \mathcal{D}(\boldsymbol{x};\frac{\partial u}{\partial \boldsymbol{x}}, \frac{\partial^2 u}{\partial \boldsymbol{x}^2},\cdots,\boldsymbol{\nu}) &= 0,\quad\forall \boldsymbol{x} \in \Omega,
    \label{eq:PDE}
\end{align}
where $\mathcal{D}$ is a differential operator, $\boldsymbol{\nu}$ is a vector of parameters, and $u(\boldsymbol{x})$ denotes the solution. Moreover, $\Omega$ is a subset of $\mathbb{R}^d$. Our goal is to learn $u(\boldsymbol{x})$ given the boundary conditions for discrete locations on the boundary as follows 
\begin{align}
    u(\boldsymbol{x}) =  g(\boldsymbol{x}), \quad\forall \boldsymbol{x}\in \partial \Omega,
    \label{eq:BC}
\end{align}
where $g(\boldsymbol{x})$ is a known function. To begin with, let us discuss supervised machine learning.
\subsection{Supervised Machine Learning}
In this section, we discuss supervised machine learning mainly for two reasons. First, supervised learning models are simpler and easier to understand than their physics-based counterparts. Therefore, we create a baseline data-driven regression model to train in a supervised learning fashion. Second, we aim to highlight the effect of scale disparity on learning when the domain objective is replaced with physics. Supervised machine learning is finding a parametric model function that maps the input data to its corresponding supervised labels \cite{murphy2012machine}. The model is trained until it can extract the embedded patterns and relationships between the pair of input-label data. The model is expected to yield accurate predictions when presented with never-before-seen data during testing. Supervised learning of the solution $u(\boldsymbol{x})$ given a set of training data $\{ \boldsymbol{x}^{(i)},u^{(i)} \}_{i=1}^{i=N_{\Omega}}$ in the domain $\Omega$ and a set of training data $\{ \boldsymbol{x}^{(i)},g^{(i)} \}_{i=1}^{i=N_{\partial \Omega}}$ on the boundary $\partial \Omega$, can be achieved by training a neural network model parameterized by $\theta$ on the following objective function
\begin{align}
    \mathcal{L}(\theta) &= \mathcal{L}_{\Omega}(\theta)  + \mathcal{L}_{\partial \Omega}(\theta),
    \label{eq:data_driven_loss}
\end{align}
where $\mathcal{L}_{\Omega}(\theta)$ is the objective function in the domain $\Omega$ 
\begin{align}
        \mathcal{L}_{\Omega}(\theta) = \frac{1}{N_{\Omega}} \sum_{i=1}^{N_{\Omega}} \| u(\boldsymbol{x}^{(i)};\theta) - u(\boldsymbol{x}^{(i)})\|_2^2,
        \label{eq:data_driven_domain_loss}
\end{align}
$\mathcal{L}_{\partial \Omega}(\theta)$ is the objective function on the boundary $\partial \Omega$ 
\begin{align}
    \mathcal{L}_{\partial \Omega}(\theta) = \frac{1}{N_{\partial \Omega}} \sum_{i=1}^{N_{\partial \Omega}} \| u(\boldsymbol{x}^{(i)};\theta) - g(\boldsymbol{x}^{(i)})\|_2^2,
\end{align}
and $\|\cdot\|_2$ is the Euclidean norm. We deliberately formulated separate objective functions for the domain and the boundary in \eqref{eq:data_driven_loss}. Later, we only change our domain objective function for our physics-informed neural network model. Hence, we encourage the readers to keep in mind this particular objective function \eqref{eq:data_driven_loss} while reading later sections. We emphasize that a vital requirement of the supervised learning method is its dependency on a vast amount of training data which may not be affordable to gather in scientific applications. Moreover, data-driven regression models are physics agnostic and cannot extrapolate. Next, we discuss a scenario where training data is unavailable, and we leverage governing equations to learn the solution $u(\boldsymbol{x})$.
\subsection{Physics-informed Neural Networks}\label{sec:PINNs}
In this section, we elaborate on the technical aspects of learning the solution $u(\boldsymbol{x})$ given a governing equation instead of training pairs of input-label data. Following the approach in \cite{dissanayake1994neural,van1995neural,Parisi2003solving,raissi2019deep}, we formulate a composite objective function similar to \eqref{eq:data_driven_loss}. Particularly, we replace the domain loss in \eqref{eq:data_driven_domain_loss} with the residuals approximated on the PDE as given in \eqref{eq:PDE}. The final objective function reads 
\begin{align}
    \mathcal{L}(\theta) = \frac{1}{N_{\Omega}} \sum_{i=1}^{N_{\Omega}}  \| \mathcal{D}(\boldsymbol{x}^{(i)};\nu)\|_2^2 +  \frac{1}{N_{\partial \Omega}} \sum_{i=1}^{N_{\partial \Omega}} \| u(\boldsymbol{x}^{(i)};\theta) - g(\boldsymbol{x}^{(i)})\|_2^2,
    \label{eq:pinn_loss}
\end{align}
where $N_{\Omega}$ is the number of collocation points in the domain $\Omega$ and $N_{\partial \Omega}$ is the number of training points on the boundary. Here, we note that the strong form of the governing equation is used in formulating the objective function \eqref{eq:pinn_loss}, which cannot allow learning of problems with non-smooth solutions. A key observation in \eqref{eq:pinn_loss} is that the objective functions for the boundary and the domain have different physical scales. It is not advisable to sum objective functions with different scales to form a mono-objective optimization equation \cite{lobato2017multi}. In \cite{basir2022critical}, we showed that scale discrepancy in a physics-informed multi-objective loss function severely impedes and may even prevent learning. A common remedy to tackle this issue is to minimize a weighted sum of multiple objective functions balanced with multiplicative weighting coefficients or hyperparameters to balance the interplay between each objective term. A weighted multi-objective function of \eqref{eq:pinn_loss} can be written as follows
\begin{align}
    \mathcal{L}(\theta) = \frac{\lambda_{\Omega}}{N_{\Omega}} \sum_{i=1}^{N_{\Omega}}  \| \mathcal{D}(\boldsymbol{x}^{(i)};\nu)\|_2^2 +  \frac{\lambda_{\partial \Omega}}{N_{\partial \Omega}} \sum_{i=1}^{N_{\partial \Omega}} \| u(\boldsymbol{x}^{(i)};\theta) - g(\boldsymbol{x}^{(i)})\|_2^2,
    \label{eq:weighted_pinn_loss}
\end{align}
where $\lambda_{\Omega}$ and $\lambda_{\partial \Omega}$ are hyperparameters that are not known a priori. Proper selection of these hyperparameters is problem-specific and significantly impacts the accuracy of model predictions. 
 A validation dataset is used for tuning hyperparameters for conventional machine-learning applications. However, when solving PDEs, the solution is the desired outcome, and no corresponding dataset is available for tuning purposes. Hence, the lack of a validation dataset and prior knowledge of the solution renders PINNs impractical for the solution of PDEs. Several heuristic methods have been proposed to balance the interplay between the objective terms in the loss function \cite{van2022optimally,wang2021understanding,liu2021dual,mcclenny2020self}. However, the solution of well-posed partial differential equations requires strict satisfaction of the boundary or initial condition and the governing PDE. In our previous work, we proposed physics $\And$ equality-constrained artificial neural networks (PECANNs) that employ the augmented Lagrangian method to constrain noiseless boundary conditions or any high fidelity data on the strong form of the governing PDE \cite{PECANN_2022}.

\section{Investigation of Failure Modes}
\label{sec:investigation_failure_modes}
\begin{figure}[!h]
\centering
    \subfloat[]{\includegraphics[scale=0.43]{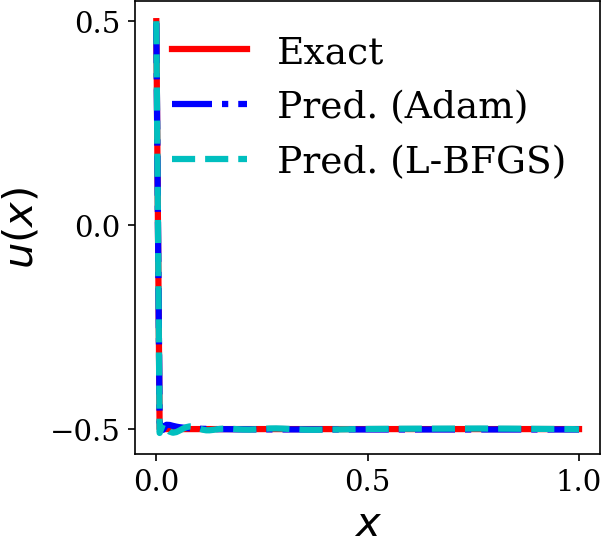}}\qquad
    \subfloat[]{\includegraphics[scale=0.43]{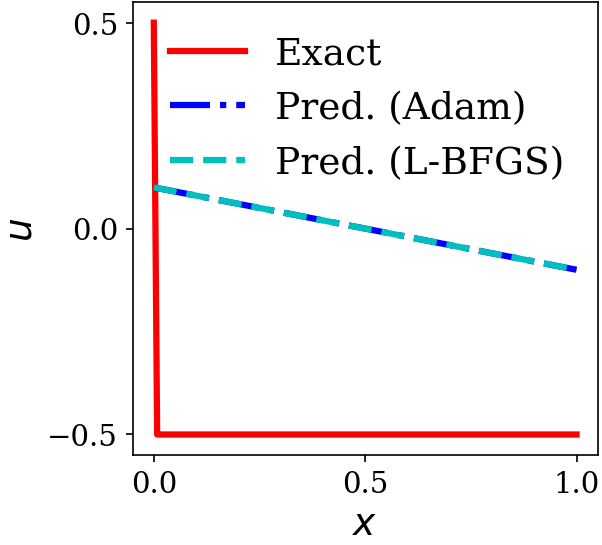}}\qquad
    \subfloat[]{\includegraphics[scale=0.43]{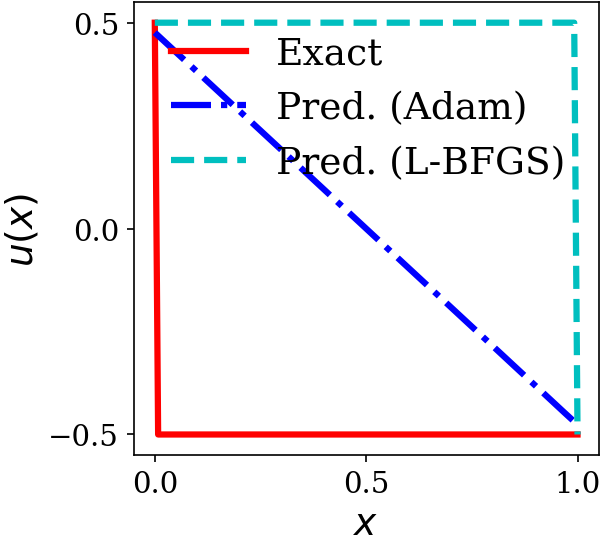}}
    \caption{Convection-dominated convection diffusion equation: predictions obtained from models trained with L-BFGS and Adam optimizer separately. The solid (red) line represents the exact solution. In contrast, the dashed(cyan) represents the predicted solution obtained by training our model with L-BFGS optimizer, and the dashed (blue) represents the predicted solution obtained by training our model with Adam optimizer. (a) data-driven model, (a) PINN model, (c) PECANN model}
    \label{fig:conv_diff_u}
\end{figure}

In the previous section, we illustrated the technical aspects of learning a target function with a supervised learning approach given a set of input-label pairs. Similarly, we discussed physics-based learning approaches that leverage the governing equations when labeled training data is unavailable. Here, we aim to demonstrate and investigate failures of existing physics-based neural network approaches in learning the solution of PDEs. Our workflow for this investigation is as follows: To begin with, we generate a set of input-target data to train a neural network model in a supervised fashion. We generate the targets from the exact solution. The key purpose for creating a data-driven model is threefold. First, we demonstrate that our neural network can represent the target function. Second, we create a baseline model to compare our physics-based models with. Third, we demonstrate that learning is successful when there is no scale disparity between the boundary and the domain objectives. We then train the same neural network on physics using PINN and PECANN frameworks separately. To reliably evaluate our models, we consider a problem with an exact solution. Thus, we consider a convection-diffusion equation that has been challenging the learn its solution with physics-based machine learning approaches. In one space dimension, the equation reads as
\begin{align}
    \frac{d u(x)}{d x} + \alpha \frac{d^2 u(x) }{dx^2} = 0, x \in (0,1),
    \label{eq:convection_diffusion_pde}
\end{align}
along with Dirichlet boundary conditions $u(0) = 0.5$ and $u(1) = -0.5$, where $u(x)$ denotes the target solution and $\alpha$ is the diffusivity coefficient. This example problem is important because decreasing $\alpha$ can result in a boundary layer in which the solution behaves drastically differently. 
Analytical solution of the above problem \cite{van2022optimally} reads as 
\begin{equation}
    u(x) = \frac{e^{-\frac{x}{\alpha}}}{1 - e^{-\frac{1}{\alpha}}} - \frac{1}{2}, \quad  x \in (0,1).
    \label{eq:convection_diffusion_exact}
\end{equation}
\citet{van2022optimally} studied this problem and reported that neither PINN nor their proposed methods produced acceptable results for $\alpha \le 10^{-3}$ with adaptive collocation sampling. We want to emphasize that in this work, we make the problem even more challenging by setting $\alpha = 10^{-6}$. We use a four-hidden layer feed-forward neural network architecture with 20 neurons per layer which is the same architecture as in \cite{van2022optimally}. We generate our collocation points on a uniform mesh to ensure that our models are trained on identical data across the domain for fair comparison purposes. After our models are trained, we visualize their loss landscapes to gain further insight into the failures of our models. We also study the impact of the choice of first-order and second-order optimizers on learning the target solution. Hence, we train our models separately with Adam\cite{kingma2014adam} optimizer and L-BFGS\cite{nocedal1980updating} optimizer. The initial learning rate for our Adam optimizer is set to $10^{-2}$, and the line search function for our L-BFGS optimizer is \emph{strong Wolfe}, which is built in PyTorch \cite{paszke2019pytorch}. We generate 2048 collocation points in the domain along with the boundary conditions only once before training. For our data-driven regression model, we obtain exact labels from our exact solution as in \eqref{eq:convection_diffusion_exact}. We train our models for 5000 epochs. We present the predictions from our models in Figure~\ref{fig:conv_diff_u}. 

Results in Figure~\ref{fig:conv_diff_u}(a) show that our baseline regression model accurately learned the underlying solution regardless of the optimizer's choice. That means our neural network architecture is capable of representing the target function. However, from Figure~\ref{fig:conv_diff_u}(b), our PINN model failed to learn the underlying target solution. We know that this failure is not due to insufficient expressivity of our neural network architecture. The only change we made to our regression objective function \eqref{eq:data_driven_loss} was replacing the domain loss with physics \eqref{eq:pinn_loss}. Consequently, our loss function was imbalanced due to scale disparity, which impeded the convergence of our optimizers \eqref{eq:pinn_loss}. Similarly, from Figure~\ref{fig:conv_diff_u}(c), our PECANN model accurately captured the boundary conditions but failed to learn the underlying solution. Again this failure is not due to the low expressivity of our neural network model. We also know that PECANNs properly balance each objective term, unlike PINNs. Thus far, we demonstrated failures of our physics-based models in learning the solution of \eqref{eq:convection_diffusion_pde}.
To gain further insight, we visualize the loss landscapes of our trained models in the next section. 

\subsection{Visualizing the Loss Landscapes}
In the previous section, we demonstrated that failures of our physics-based models were not due to the insufficient expressivity of our neural network architecture. Moreover, the choice of an optimizer did not impact the predictions of our physics-based models in learning the solution of \eqref{eq:convection_diffusion_pde}. In this section, we visualize the loss landscapes of our trained models to help understand why our models failed to learn the underlying target solution when trained on physics instead of labeled data. The loss landscape provides insights into the geometry of the optimization problem, including the presence of saddle points, the number of local minima, and the geometric shape of the valleys that represent solutions. These insights can help to explain why some optimization algorithms converge to reasonable solutions while others get stuck in poor local minima. 
\citet{li2018visualizing} demonstrated that simple visualization methods could fail to capture the local geometry of loss function minimizers, leading to a poor understanding of the optimization problem. To address this issue, the authors proposed a filter normalization technique that better illustrates the relationship between the sharpness of minima and the model's generalization error. We should be cautious that non-convexity in a reduced dimensional plot implies that non-convexity exists in the full-dimensional surface. However,  the appearance of convexity in a low-dimensional representation does not guarantee that the high-dimensional function is convex \cite{li2018visualizing}. In this work, we use the 'filtered normalization' technique to visualize our loss landscapes. Hence, we choose a center $\theta^*$ and two filter-wise normalized random vectors $\zeta$ and $\gamma$ \cite{li2018visualizing}. We then plot a function of the form
\begin{align}
    f(\epsilon_1, \epsilon_2) = \mathcal{L}(\theta^*+ \epsilon_1 \zeta + \epsilon_2  \gamma)
\end{align}
in log scale, where $\mathcal{L}$ is the objective function of our respective neural network model (i.e., regression model, PINN or PECANN model), $\epsilon_1 \in [-1,1] $ and $\epsilon_2 \in [-1,1] $ are steps in the direction of each random vectors respectively. It is worth noting that $\theta^*$ is the state of our trained models. Now, let us visualize the loss landscapes of our trained models. 
\begin{figure}[!ht]
 \centering
    \subfloat[]{\includegraphics[scale=0.45]{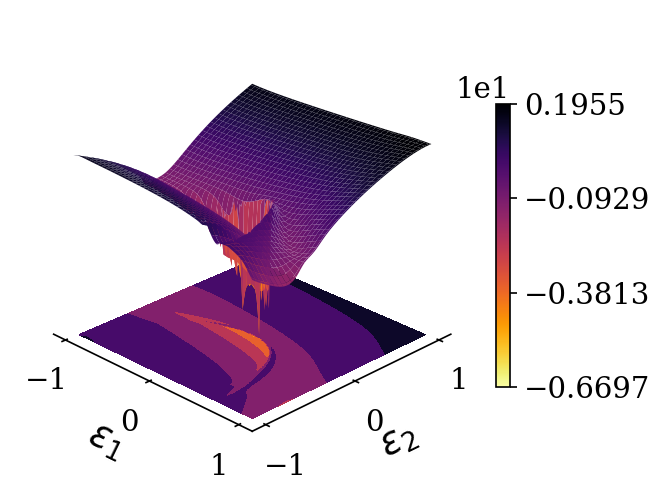}}
    \subfloat[]{\includegraphics[scale=0.45]{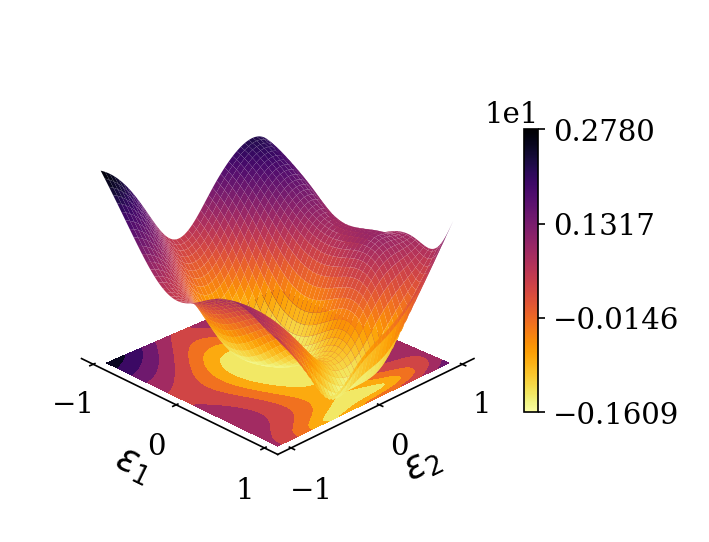}}
    \subfloat[]{\includegraphics[scale=0.45]{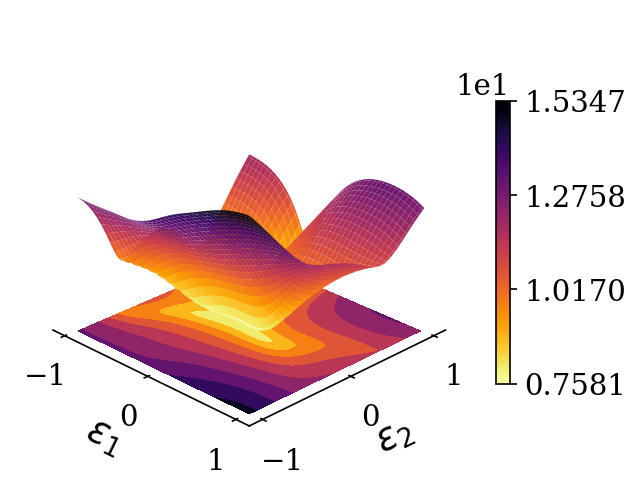}}
    \\
    \subfloat[]{\includegraphics[scale=0.45]{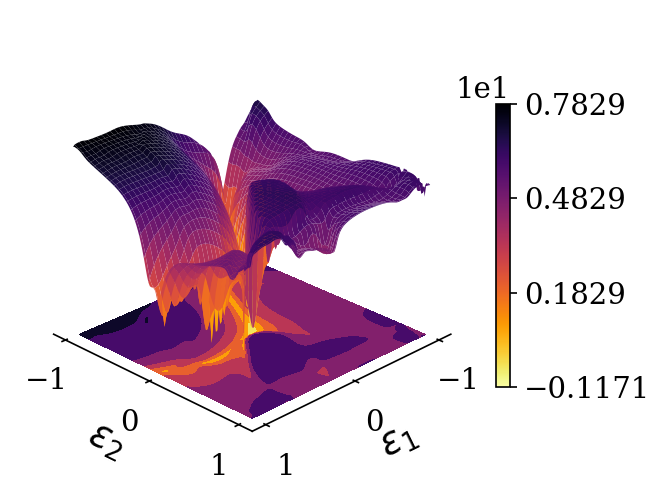}}
    \subfloat[]{\includegraphics[scale=0.45]{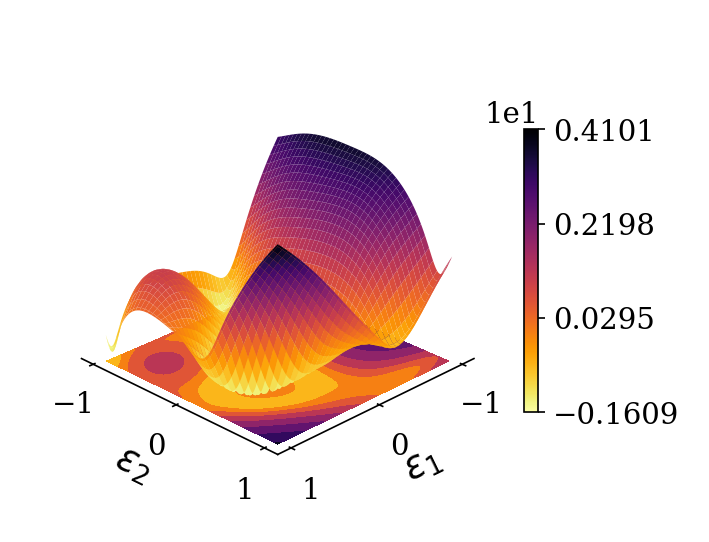}}
    \subfloat[]{\includegraphics[scale=0.45]{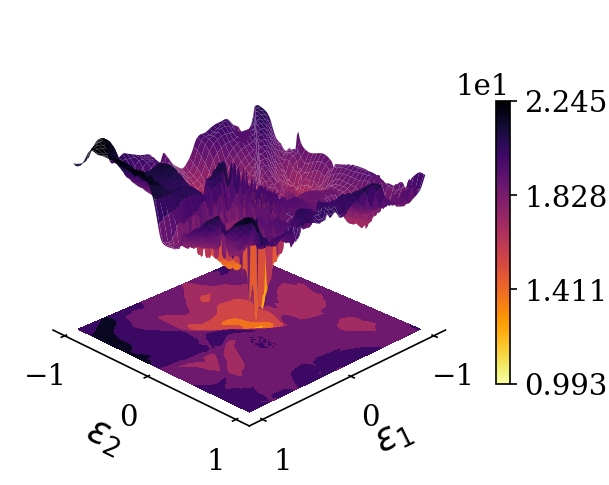}}
    \caption{Loss landscapes of different models trained with Adam and L-BFGS optimizers. Top row: Models trained with Adam optimizer. (a) data-driven model,(b) PINN model, (c) PECANN model. Bottom row: models trained with L-BFGS optimizer. (d) data-driven model, (e) PINN model, (f) PECANN model. }
    \label{fig:conv_diff_loss_landscapes}
\end{figure}

From Figure~\ref{fig:conv_diff_loss_landscapes}(a), we observe that our baseline regression model produced smooth landscapes and converged to a good local minimum when trained with Adam optimizer. Nevertheless, the two-dimensional loss landscapes exhibit a rough and uneven terrain, with a distinct minimum that our model converges to when trained with L-BFGS optimizer, as illustrated in Figure~\ref{fig:conv_diff_loss_landscapes}(d). That is because L-BFGS optimizer incorporates the curvature information of the objective function into the training process. To this end, we have seen that regardless of the choice of the optimizer, our purely data-driven regression model accurately learned the underlying solution and converged to a good local minimum. In contrast to the baseline regression model, our PINN model converged to a saddle point with a flat region around it when trained with Adam optimizer, as depicted in Figure~\ref{fig:conv_diff_loss_landscapes}(b). Similarly, in Figure~\ref{fig:conv_diff_loss_landscapes}(e), we observe that the low dimensional landscape of our PINN model is non-convex when trained with L-BFGS optimizer. Similarly, our PECANN model converged to a flat basin of attraction when trained with Adam optimizer, as shown in Figure~\ref{fig:conv_diff_loss_landscapes}(c). Finally, our PECANN model produces highly uneven loss landscapes when trained with L-BFGS optimizer as shown in Figure~\ref{fig:conv_diff_loss_landscapes}(f).
So far, we have observed and characterized the difficulty of navigating loss landscapes produced by training a neural network model on physics. Contrary to our expectations, L-BFGS optimizer failed to effectively traverse the loss landscapes generated by our physics-based neural network models, despite utilizing the curvature information of the objective function during optimization. We aim to answer why existing physics-based models produced highly complex and non-convex loss landscapes.

\subsection{Sensitivity Analysis of Backpropagated Gradients in the Presence of Physics}
\label{sec:effect_of_Diff_Opt}
In the previous section, we visualized the loss landscapes of our trained models and observed that models trained on physics can get stuck in bad local minimums regardless of the choice of an optimizer. We re-emphasize that two-dimensional slices of loss landscapes could be misleading and unintuitive. In this section, we explore an alternative approach to find why existing physics-based neural network models produce highly complex loss landscapes. In particular, we demonstrate how a differential operator pollutes the back-propagated gradients, the backbone of training artificial neural networks. 

\begin{figure}[!ht]
 \centering
    \subfloat[]{\includegraphics[scale=0.45]{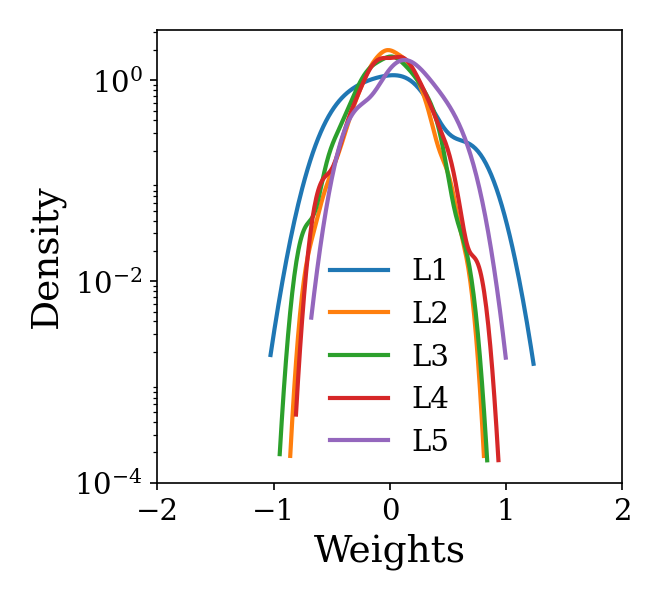}}
    \subfloat[]{\includegraphics[scale=0.45]{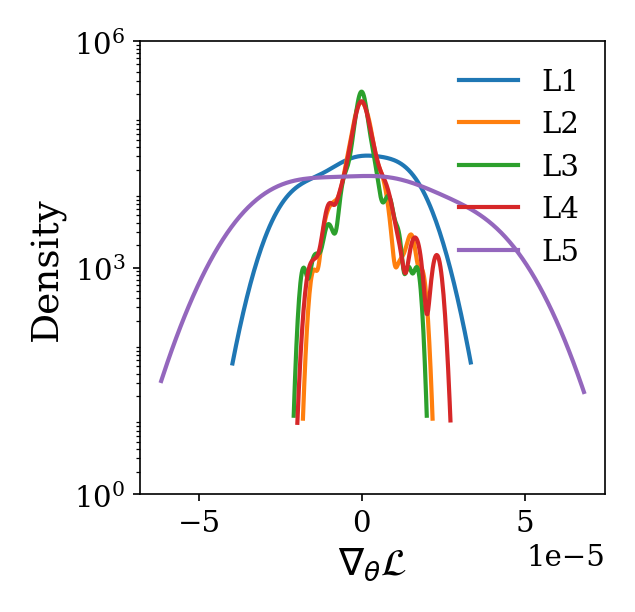}}
    \subfloat[]{\includegraphics[scale=0.45]{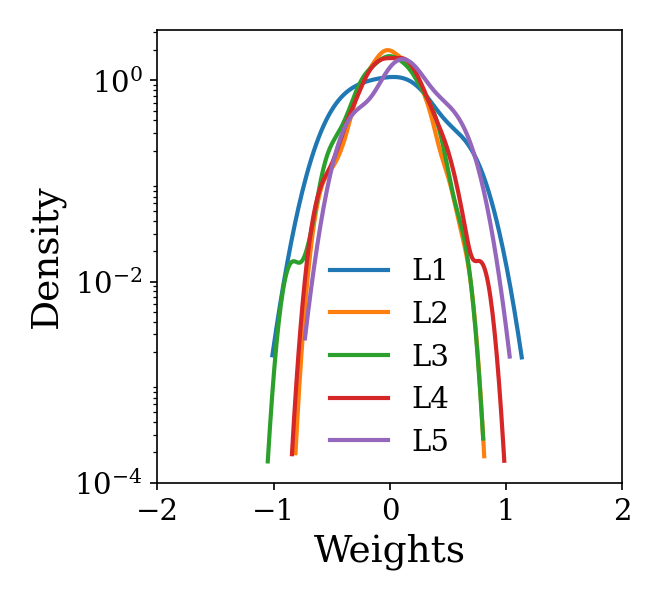}}\\
    \subfloat[]{\includegraphics[scale=0.45]{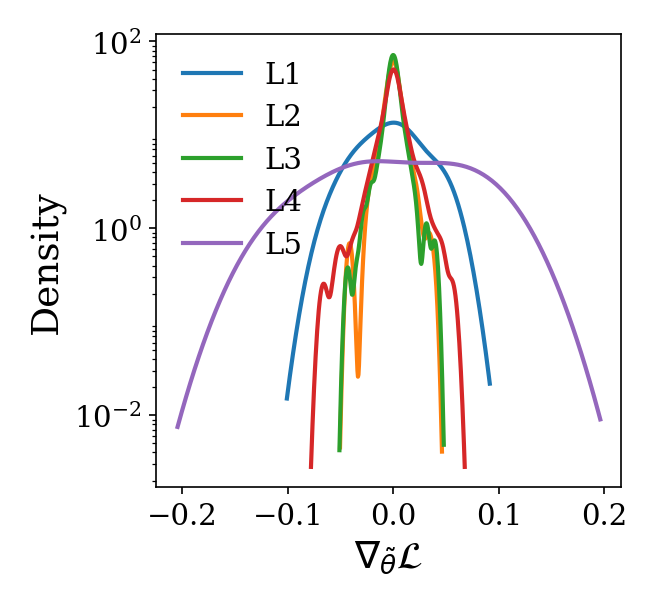}}\quad
    \subfloat[]{\includegraphics[scale=0.45]{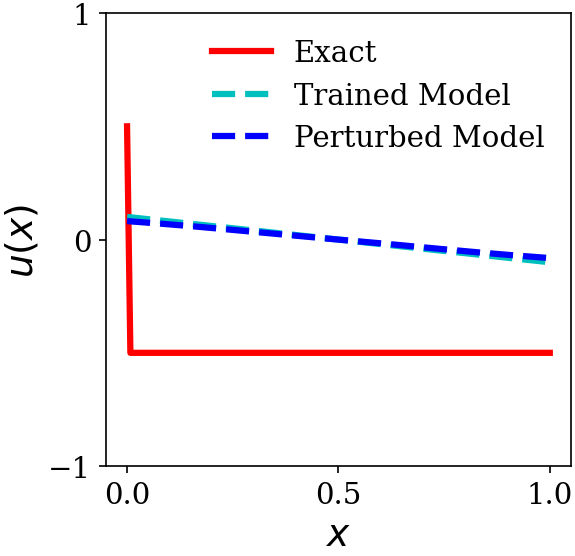}}\quad
    \subfloat[]{\includegraphics[scale=0.45]{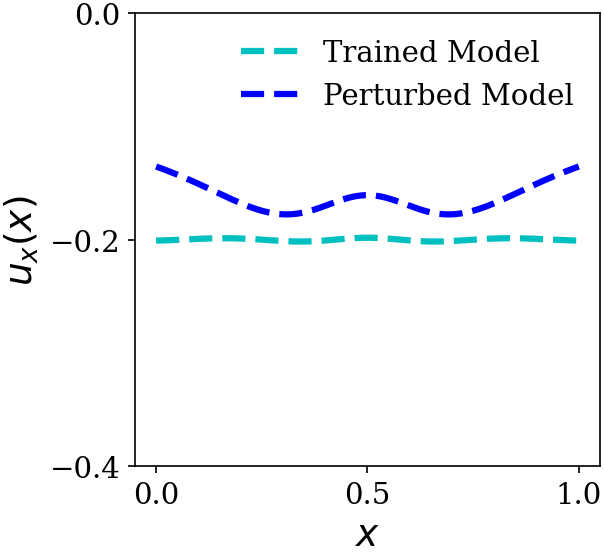}}
    \caption{effect of a differential operator in training our PINN model with L-BFGS optimizer: (a) distribution of parameters of the network before perturbation, (b) distribution of gradients of the parameters of the network before perturbation, (c) distribution of parameters of the network at the perturbed state, (d) distribution of gradients of the parameters of the network at the perturbed state, (e) prediction before and after perturbation (f) predicted derivative before and after perturbation}
    \label{fig:noise_amplification_pinn_model}
\end{figure}

Neural networks are initialized randomly before training\cite{glorot2010understanding,he2015delving}. As a result, predictions obtained from a neural network model are random and extremely noisy, particularly during the early stages of training. Therefore, a differential operator applied to noisy predictions leads to incorrect derivatives. As a result, physics loss becomes extremely noisy or incorrect. Taking the gradient of this noisy physics loss results in highly incorrect backpropagated gradients. Essentially, severely contaminated gradients produce undesirable consequences during training since backpropagated gradients are the backbone of training artificial neural networks. This issue becomes severe for high-order differential operators. We also conjecture that pre-training and transfer learning are effective because the model is initialized from a better state that produces less noisy predictions from a neural network model, particularly during the early stages of training. We leave that discussion to our future work as it is beyond the scope of the current paper. To quantify the impact of a differential operator in corrupting the backpropagated gradients, we pursue the following approach. First, we record the current state of our model (i.e., parameters and gradient of parameters). We then slightly perturb the state of our model as follows, 
\begin{align}
    \mathcal{N}(\boldsymbol{x};\Tilde{\theta}) =\mathcal{N}(\boldsymbol{x};\theta^* + \epsilon_1 \zeta + \epsilon_2 \gamma),
\end{align}
where $\mathcal{N}$ is our neural network model, $\theta^*$ is the undisturbed state of the model, $\Tilde{\theta}$ is the perturbed state of our model, $\zeta$ and $\gamma$ are filter-normalized random vectors\cite{li2018visualizing} and $\epsilon$ is a small positive number. Next, we make predictions from our model at the perturbed state $\mathcal{N}(\boldsymbol{x};\Tilde{\theta})$ and calculate our loss. Finally, we calculate the gradients of our parameters with respect to our loss. The key point is that we can compare the backpropagated gradients before and after the perturbation. The result of our analysis for our PINN model trained with L-BFGS optimizer is reported in Figure~\ref{fig:noise_amplification_pinn_model}.

\begin{figure}[!ht]
 \centering
    \subfloat[]{\includegraphics[scale=0.45]{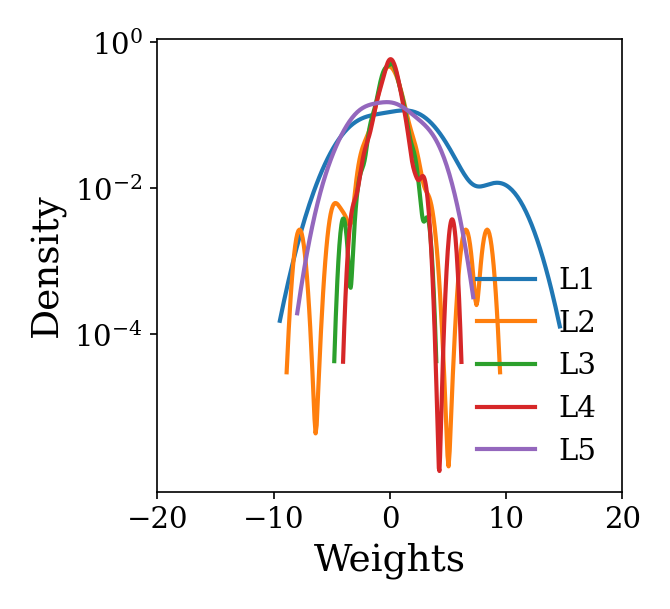}}
    \subfloat[]{\includegraphics[scale=0.45]{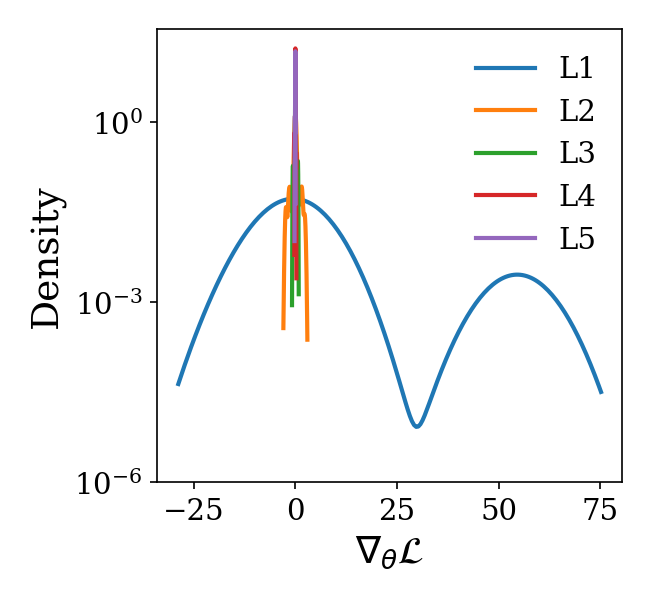}}
    \subfloat[]{\includegraphics[scale=0.45]{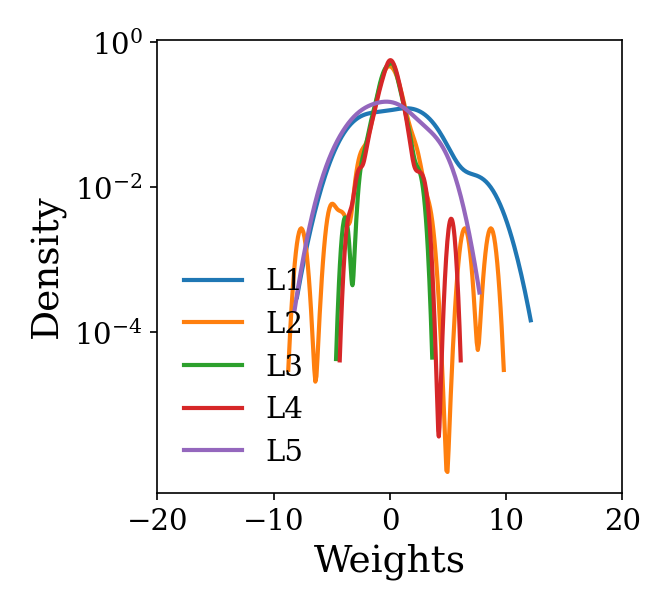}}\\
    \subfloat[]{\includegraphics[scale=0.45]{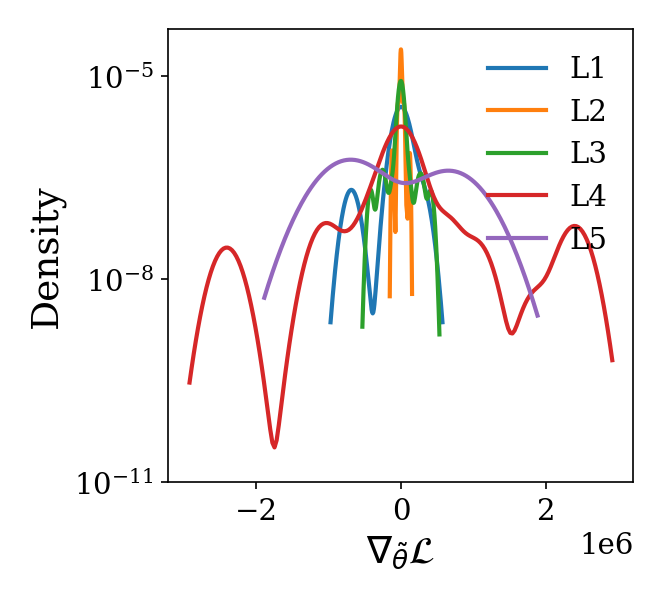}}\quad
    \subfloat[]{\includegraphics[scale=0.45]{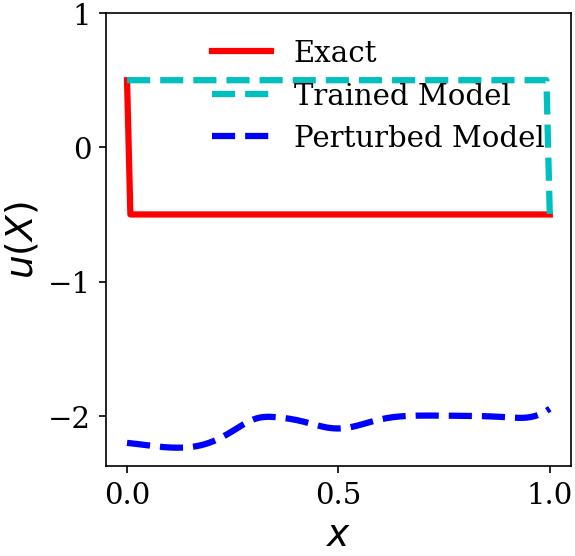}}\quad
    \subfloat[]{\includegraphics[scale=0.45]{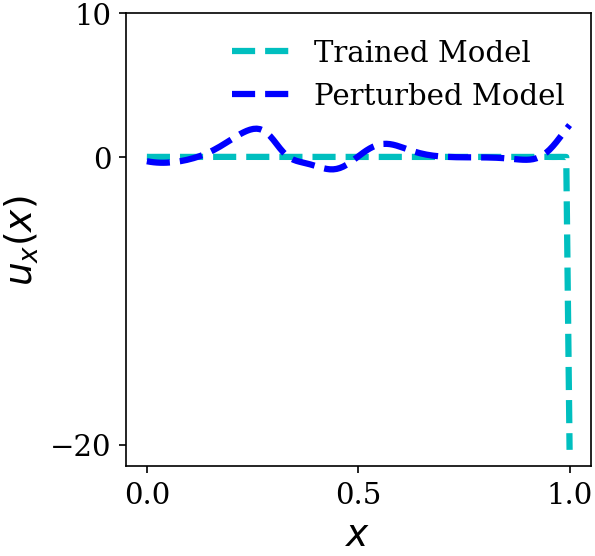}}
    \caption{effect of a differential operator in training our PECANN model with L-BFGS optimizer:(a) distribution of parameters of the network before perturbation, (b) distribution of gradients of the parameters of the network before perturbation, (c) distribution of parameters of the network at the perturbed state, (d) distribution of gradients of the parameters of the network at the perturbed state,(e) prediction before and after perturbation (f) predicted derivative before and after perturbation}
    \label{fig:noise_amplification_pecann_model}
\end{figure}

From Figures~\ref{fig:noise_amplification_pinn_model}(a)-(c), we observe that perturbations are small since the distribution of the parameters before and after perturbations are similar. However, backpropagated gradients have increased by almost five orders of magnitude, as can be seen from Figures~\ref{fig:noise_amplification_pinn_model}(b)-(d). We also observe the impact of perturbations in predictions of our PINN model in Figure~\ref{fig:noise_amplification_pinn_model}(e). However, a small perturbation in the prediction produced an entirely different derivative, as seen in Figure~\ref{fig:noise_amplification_pinn_model}(f). A similar analysis of our PINN model trained with Adam optimizer is provided in the appendix \ref{sec:appendix_sensitivity_gradients}. Similarly, we report the summary of our sensitivity analysis of backpropagated gradients for our PECANN model trained with L-BFGS optimizer in Figure~\ref{fig:noise_amplification_pecann_model}.

 The perturbation injected in our PECANN model is small, as can be seen from the distribution of the undisturbed and perturbed parameters before and after perturbations are similar in Figures~\ref{fig:noise_amplification_pecann_model}(a)-(c). However, this small perturbation severely contaminated the backpropagated gradients, as can be seen in Figures~\ref{fig:noise_amplification_pecann_model}(b)-(d). The consequence of this perturbation on the prediction and the derivative is markedly visible in Figures~\ref{fig:noise_amplification_pecann_model}(e)-(f). A similar analysis of our PECANN model trained with Adam optimizer is provided in appendix \ref{sec:appendix_sensitivity_gradients}. In a nutshell, we observed that differential operators amplify the embedded noise in the predictions of a physics-based neural network model. This issue becomes severe for high-order differential operators.
As a consequence, backpropagated gradients get contaminated, which prevents convergence. This is why we propose preconditioning differential operators by introducing auxiliary variables. Next, we discuss our proposed formulation, constrained optimization problem, and our unconstrained dual problem.
\section{Proposed Method}\label{sec:Proposed_Method}
In this section, we illustrate the key ingredients of our proposed method. Without loss of generality, let us start considering Stokes's equation,
\begin{subequations}
\begin{align}
    \nabla \cdot \boldsymbol{u}(\boldsymbol{x}) &=0,\quad \text{in} \quad \Omega \\
    \frac{1}{Re}\Delta \boldsymbol{u}(\boldsymbol{x}) + \nabla p(\boldsymbol{x}) &=\boldsymbol{f}(\boldsymbol{x}), \quad \text{in} \quad \Omega\\
    \boldsymbol{u}(\boldsymbol{x}) &=0, \quad \text{on} \quad \partial \Omega
\end{align}
\end{subequations}
where $p$ is the pressure, $\boldsymbol{u}$ is the velocity vector, $\boldsymbol{f}$ is the body force vector, $\Delta$ is the Laplacian operator, and $Re$ is the Reynolds number. Stokes equation has many applications in engineering and physics. Stokes operator is the primary ingredient of Navier–Stokes equations that can describe the physics of various phenomena in science and engineering. \citet{bo1990least} proposed a least-squares finite element method based on a first-order velocity-pressure-vorticity formulation that can handle multi-dimensional and incompressible problems Navier-Stokes equations. In this work, we adopt the approach in \cite{bo1990least} and introduce an auxiliary vorticity variable $\boldsymbol{\omega} = \nabla \times \boldsymbol{u}$ to reduce the above problem to a system of first-order differential equations written in its residual form as follows
\begin{subequations}
\begin{align}
     \mathcal{C}(\boldsymbol{x}) &=\nabla \cdot \boldsymbol{u}(\boldsymbol{x}) ,\quad \text{in} \quad \Omega \\
     \mathcal{D}(\boldsymbol{x}) &= \frac{1}{Re}\nabla \times \boldsymbol{\omega}(\boldsymbol{x}) + \nabla p(\boldsymbol{x}) - \boldsymbol{f}(\boldsymbol{x}), \quad \text{in} \quad \Omega\\
    \mathcal{F}(\boldsymbol{x}) &=  \boldsymbol{\omega}(\boldsymbol{x}) - \nabla \times \boldsymbol{u}(\boldsymbol{x}),\quad \text{in} \quad \Omega\\
    \mathcal{B}(\boldsymbol{x}) &= \boldsymbol{u}(\boldsymbol{x}) - 0, \quad \text{on} \quad \partial \Omega.
\end{align}
\label{eq:reduced_order_pde}
\end{subequations}

The critical point here is that by reducing the order of a partial differential equation, we bypass the need to calculate high-order derivatives. This reduces the search space of our solution and facilitates learning problems with non-smooth solutions that we will show in our numerical experiments. Moreover, first-order differential equations are easier to learn than high-order differential operators due to several issues we discussed in previous sections. Introducing an auxiliary vorticity variable is not the only way to reduce the order of a partial differential equation. We can also introduce other variables (i.e., auxiliary flux variable) to reduce the order of a given problem, as we will show in our numerical experiments. It is worth keeping in mind that our method is meshless and geometry invariant. Next, we discuss our constrained optimization formulation.

\subsection{Constrained Optimization Problem Formulation}

Here, we discuss the key ingredients of our constrained optimization formulation. The objective is to learn the underlying solution $u(\boldsymbol{x})$ , $p(\boldsymbol{x})$ and $\boldsymbol{\omega}(\boldsymbol{x})$ for the problem given in \eqref{eq:reduced_order_pde}. Therefore, we create a neural network model $\mathcal{N}(\boldsymbol{x};\theta): \boldsymbol{x} \rightarrow [u(\boldsymbol{x}),p(\boldsymbol{x}),\boldsymbol{\omega}(\boldsymbol{x})]$ that takes $\boldsymbol{x}$ as inputs and predicts $u(\boldsymbol{x}),p(\boldsymbol{x})$ and $\boldsymbol{\omega}(\boldsymbol{x})$. It is worth noting that we use a single neural network architecture to predict our primary variables and our auxiliary variable instead of two separate neural network architectures. That is to reduce the risk of overfitting and to encourage learning common hidden representations for our primary and auxiliary variables. Given a set of collocation points $\{\boldsymbol{x}^{(i)} \}_{i=1}^{N_{\Omega}}$ sampled in the domain $\Omega$, and a set of boundary points $\{\boldsymbol{x}^{(i)} \}_{i=1}^{N_{\partial \Omega}}$ sampled on the boundary $\partial \Omega$, we minimize
\begin{equation}
    \min_{\theta} \quad \mathcal{J}_{\mathcal{D}}(\theta) = \sum_{i=1}^{N_{\Omega}} \|\mathcal{D}(\boldsymbol{x}^{(i)},\theta)\|^2_2
\end{equation}
subject to the following constraints 
\begin{subequations}
\begin{align}
    -\epsilon &\le \mathcal{B}(\boldsymbol{x}^{(i)};\theta) \le \epsilon, \quad \forall i = 1, \cdots, N_{\partial\Omega},\\
    -\epsilon &\le \mathcal{C}(\boldsymbol{x}^{(i)};\theta) \le \epsilon, \quad \forall i = 1, \cdots, N_{\Omega},\\
    -\epsilon &\le \mathcal{F}(\boldsymbol{x}^{(i)};\theta) \le \epsilon, \quad \forall i = 1, \cdots, N_{\Omega},
\end{align}
\label{eq:constraints}
\end{subequations}

Where $\epsilon > 0 $ is a small positive number. We want our constraint interval to be small to ensure the errors are as small as possible. It is also important to note that we strictly enforce the boundary conditions $\mathcal{B}$. Moreover, we strictly enforce the compatibility equation between our primary variable $u$ and our auxiliary variable $\boldsymbol{\omega}$ as given in $\mathcal{F}$. In addition, we precisely enforce our primary variable $u$ to be divergence-free as given in $\mathcal{C}$. As a result of our constraints, our model locally conserves the laws and the boundary conditions. 

One can make the argument to aggregate the loss on $\mathcal{F}$, $\mathcal{C}$, and $\mathcal{D}$ since they are all first-order differential equations. However, these equations have different scales which cannot be aggregated together. Another important point is that constraining $\mathcal{F}$ and $\mathcal{C}$ allows our model to focus on challenging regions to learn adaptively. Let us simplify our constraints by introducing a convex distance function $\phi \in [0,\infty)$(i.e., quadratic function) and setting $\epsilon = 0$. We do this step because we only care about the magnitude of the error, not the sign of it. Therefore, we rewrite our constraints as follows
\begin{subequations}
\begin{align}
    \phi(\mathcal{B}(\boldsymbol{x}^{(i)};\theta)) &= 0, \quad \forall i = 1, \cdots,  N_{\partial \Omega},\\
   \phi(\mathcal{C}(\boldsymbol{x}^{(i)};\theta)) &= 0, \quad \forall i = 1, \cdots,  N_{\Omega},\\
   \phi(\mathcal{F}(\boldsymbol{x}^{(i)};\theta))&= 0, \quad \forall i = 1, \cdots,  N_{\Omega}.
\end{align}
\label{eq:simplified_constraints}
\end{subequations}
It is worth noting that noisy measurement data can be seamlessly incorporated into our objective function by minimizing the log-likelihood of the predictions obtained from a neural network model conditioned on the observations \cite{PECANN_2022}. Next, we formulate a dual optimization problem that can be used as an objective function for training our neural network model.

\subsection{Unconstrained Optimization Problem Formulation}
\label{sec:unconstrained_formuluation}
In this section, we formulate a dual unconstrained optimization problem using Lagrange multiplier methods as follows
\begin{subequations}
\begin{align}
\min_{\theta} \max_{\lambda_{\mathcal{F}},\lambda_{\mathcal{B}},\lambda_{\mathcal{C}}}
\mathcal{L}(\theta;\lambda_{\mathcal{C}},\lambda_{\mathcal{B}},\lambda_{\mathcal{N}}) &=  \mathcal{J}_{\mathcal{D}}(\theta) + \mathcal{J}_{\mathcal{F}}(\lambda_{\mathcal{F}},\theta)+ \mathcal{J}_{\mathcal{C}}(\lambda_{\mathcal{C}},\theta)+ \mathcal{J}_{\mathcal{B}}(\lambda_{\mathcal{B}},\theta)
\\
\mathcal{J}_{\mathcal{F}}(\lambda_{\mathcal{F}},\theta) &=\sum_{i=1}^{N_{\Omega}}\lambda_{\mathcal{F}}^{(i)} \phi(\mathcal{F}(\boldsymbol{x}^{(i)};\theta)),\\
\mathcal{J}_{\mathcal{C}}(\lambda_{\mathcal{C}},\theta)
&=\sum_{i=1}^{N_{\Omega}}\lambda_{\mathcal{C}}^{(i)} \phi(\mathcal{C}(\boldsymbol{x}^{(i)};\theta)),\\
\mathcal{J}_{\mathcal{B}}(\lambda_{\mathcal{B}},\theta) &=\sum_{i=1}^{N_{\Omega}}\lambda_{\mathcal{B}}^{(i)} \phi(\mathcal{B}(\boldsymbol{x}^{(i)};\theta)).
\label{eq:dual_problem}
\end{align}
\end{subequations}
We can swap the order of the minimum and the maximum by using the following \textit{minimax} inequality concept or weak duality
\begin{subequations}
\begin{align}
   \max_{\lambda_{\mathcal{F}},\lambda_{\mathcal{B}},\lambda_{\mathcal{C}}}\min_{\theta} \mathcal{L}(\theta;\lambda_{\mathcal{F}},\lambda_{\mathcal{B}},\lambda_{\mathcal{C}})  \le \min_{\theta} \max_{\lambda_{\mathcal{F}},\lambda_{\mathcal{B}},\lambda_{\mathcal{C}}}\mathcal{L}(\theta;\lambda_{\mathcal{F}},\lambda_{\mathcal{B}},\lambda_{\mathcal{C}}).
\end{align}
\end{subequations}
Therefore, our final dual problem is given as follows 
\begin{align}
 \max_{\lambda_{\mathcal{F}},\lambda_{\mathcal{B}},\lambda_{\mathcal{C}}} \min_{\theta}
\mathcal{L}(\theta;\lambda_{\mathcal{F}},\lambda_{\mathcal{B}},\lambda_{\mathcal{C}}).
\label{eq:max_min_dual}
\end{align}

We observe that the inner part of \eqref{eq:max_min_dual} is the dual objective function which is concave even though $\mathcal{J}_{\mathcal{F}}$, $\mathcal{J}_{\mathcal{B}}$, and $\mathcal{J}_{\mathcal{C}}$ can be non-convex. The minimization can be performed using any gradient descent-type optimizer that updates the parameters $\theta$ as follows
\begin{align}
    \theta \leftarrow \theta - \alpha (H(\theta))^{-1} \nabla_{\theta} \mathcal{L}(\theta;\lambda_{\mathcal{F}},\lambda_{\mathcal{B}},\lambda_{\mathcal{C}})
\end{align}
where $\alpha$ is the learning rate and $H$ is the hessian matrix. Similarly, we can use the gradient ascent rule to update our Lagrange multipliers to perform the maximization as follows 
\begin{subequations}
\begin{align}
\lambda_{\mathcal{B}}^{(i)} &\leftarrow \lambda_{\mathcal{B}}^{(i)} + \eta \phi(\mathcal{B}(\boldsymbol{x}^{(i)};\theta)) , \forall i = 1,\cdots, N_{\partial \Omega},\\
  \lambda_{\mathcal{F}}^{(i)} &\leftarrow \lambda_{\mathcal{F}}^{(i)} + \eta \phi(\mathcal{F}(\boldsymbol{x}^{(i)};\theta)) , \forall i = 1,\cdots, N_{\Omega},\\
\lambda_{\mathcal{C}}^{(i)} &\leftarrow \lambda_{\mathcal{C}}^{(i)} +  \eta \phi(\mathcal{C}(\boldsymbol{x}^{(i)};\theta)) , \forall i = 1,\cdots, N_{\Omega},
\end{align}
\end{subequations}
where  $\xleftarrow{}$ indicates an optimization step and $\eta > 0$ is a small learning rate. To accelerate the convergence of our dual problem, we adapt the learning rate to Lagrange multipliers following the idea from adaptive subgradient methods\cite{ruder2016overview}. We should note that it is possible to adapt the learning rate $\eta$ following the approach in Adam \cite{kingma2014adam}. However, since Lagrange multipliers' gradients do not oscillate, there is no need to employ momentum acceleration. Therefore, we chose to adapt learning rate $\eta$ following the idea of RMSprop, which divides the learning rate by an exponentially decaying average of squared gradients as follows,
\begin{subequations}
\begin{align}
    \mathbb{E}[b^2(\boldsymbol{x}^{(i)})] &\leftarrow \beta \mathbb{E}[b^2(\boldsymbol{x}^{(i)})] + (1- \beta) [\phi(\mathcal{B}(\boldsymbol{x}^{(i)};\theta))]^2,\forall i = 1,\cdots, N_{\partial \Omega},\\
    \mathbb{E}[f^2(\boldsymbol{x}^{(i)})] &\leftarrow \beta \mathbb{E}[f^2(\boldsymbol{x}^{(i)})] + (1- \beta) [\phi(\mathcal{F}(\boldsymbol{x}^{(i)};\theta))]^2,\forall i = 1,\cdots, N_{\Omega}\\
    \mathbb{E}[c^2(\boldsymbol{x}^{(i)})] &\leftarrow \beta \mathbb{E}[c^2(\boldsymbol{x}^{(i)})] + (1- \beta) [\phi(\mathcal{C}(\boldsymbol{x}^{(i)};\theta))]^2,\forall i = 1,\cdots, N_{\Omega}
\end{align}
\end{subequations}
where $\mathbb{E}[\cdot]$ denotes the average operator, $\beta \in (0,1)$ indicates how important the current observation is, and $\epsilon$ is the smoothing term that avoids division by zero, which is set to $10^{-10}$ unless specified otherwise. In this work we set $\beta = 0.9$ and $\eta = 10^{-2}$ unless specified otherwise. Having calculated the exponentially decaying average of squared gradients, we can update our Lagrange multipliers using the following update rule
\begin{subequations}
\begin{align}
    \lambda_{\mathcal{B}}^{(i)} &\leftarrow \lambda_{\mathcal{B}}^{(i)} + \frac{\eta}{\sqrt{\mathbb{E}[b^2(\boldsymbol{x}^{(i)})] + \epsilon}} \phi(\mathcal{B}(\boldsymbol{x}^{(i)};\theta)) , \forall i = 1,\cdots, N_{\partial \Omega},\\
    \lambda_{\mathcal{F}}^{(i)} &\leftarrow \lambda_{\mathcal{F}}^{(i)} + \frac{\eta}{\sqrt{\mathbb{E}[f^2(\boldsymbol{x}^{(i)})] + \epsilon}} \phi(\mathcal{F}(\boldsymbol{x}^{(i)};\theta)) , \forall i = 1,\cdots, N_{\Omega}\\
     \lambda_{\mathcal{C}}^{(i)} &\leftarrow \lambda_{\mathcal{C}}^{(i)} + \frac{\eta}{\sqrt{\mathbb{E}[c^2(\boldsymbol{x}^{(i)})] + \epsilon}} \phi(\mathcal{C}(\boldsymbol{x}^{(i)};\theta)) , \forall i = 1,\cdots, N_{\Omega}.
\end{align}
\end{subequations}
We note that the magnitude of a Lagrange multiplier for a particular constraint indicates our optimizer's challenge in enforcing that constraint. Therefore, our constraints (i.e., $\mathcal{F}$ and $\mathcal{C}$) in the domain help our model focus on regions that are complex to learn. We will demonstrate this in our numerical experiments. Our dual unconstrained optimization formulation adaptively enforces boundary conditions, flux, or any user-defined equality constraints on the model prediction, eliminating the need for manual tuning or estimation. We note that our unconstrained optimization formulation is a general approach that can handle constrained optimization problems involving PDEs, as we demonstrate in a numerical example in \ref{sec:Helmholtz}. 
\section{Numerical Experiments}\label{sec:Numerical_Experiments}
We adopt the following metrics for evaluating the prediction of our models. Given an $n$-dimensional vector of predictions $\boldsymbol{\hat{u}} \in \mathbf{R}^n$ and an $n$-dimensional vector of exact values $\boldsymbol{u} \in \mathbf{R}^n$, we define a relative Euclidean or $L^2$ norm
\begin{align}
     \mathcal{E}_r(\hat{u},u) = \frac{\|\hat{\boldsymbol{u}} - \boldsymbol{u}\|_2}{\|\boldsymbol{u}\|_2}, 
    \label{eq:relativeL2Error} \quad
    \mathcal{E}_{\infty}(\hat{u},u) = \| \boldsymbol{\hat{u}} - \boldsymbol{u}\|_{\infty}, \quad
    \text{MAE} = \frac{1}{n}\sum_{i=1}^{n}(\hat{\boldsymbol{u}}^{(i)} - \boldsymbol{u}^{(i)})^2
\end{align}
where $\|\cdot \|_2$ denotes the Euclidean norm and $\| \cdot \|_{\infty}$ denotes the maximum norm. All the codes accompanying this manuscript are open-sourced at \cite{Basir_Investigating_and_Mitigating_2022}.

\subsection{Convection-dominated Convection Diffusion Equation}
\begin{figure}[!ht]
\centering
    \subfloat[]{\includegraphics[scale=0.42]{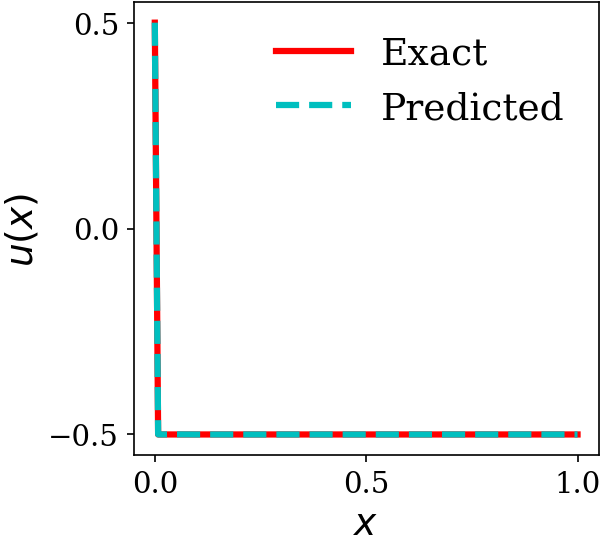}}\quad
    \subfloat[]{\includegraphics[scale=0.42]{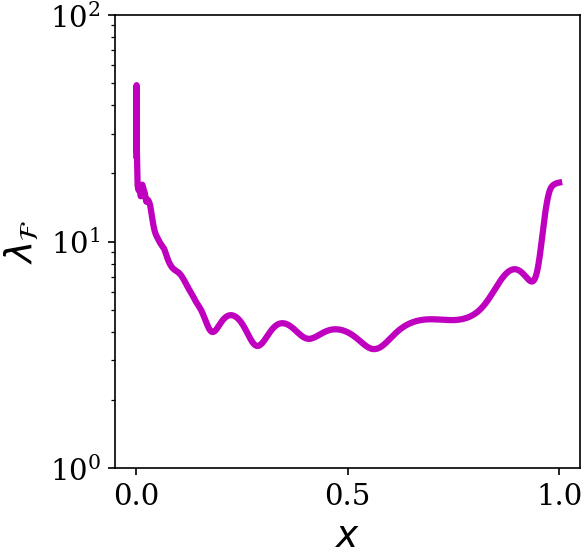}}\quad
    \subfloat[]{\includegraphics[scale=0.50]{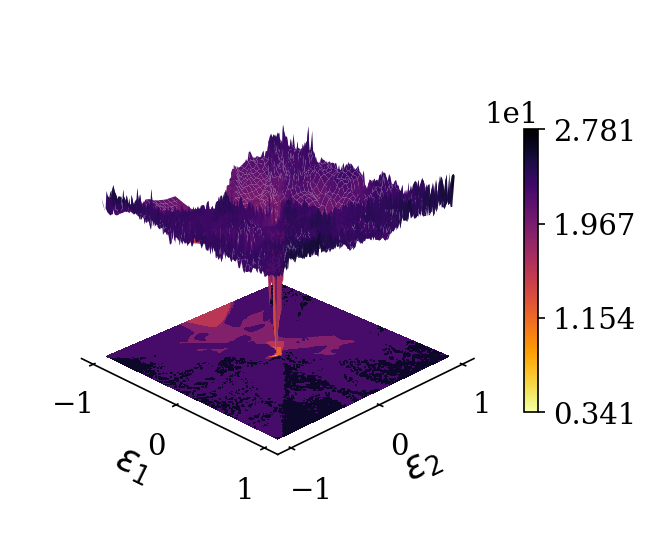}}
    \caption{Convection-dominated convection-diffusion equation with $\alpha = 10^{-6}$: (a) exact vs. predicted solution with $\mathcal{E}_r = 3.11 \times 10^{-5}$ and $\mathcal{E}_\infty = 4.75\times 10^{-5}$, (b) Lagrange multipliers for enforcing the auxiliary flux equation, (c) loss landscapes of our trained model.}
    \label{fig:proposed_convection_diffusion}
\end{figure}

In this section, we aim to learn the solution for a convection-dominate convection-diffusion equation given in \eqref{eq:convection_diffusion_pde} along with its boundary conditions. In section \ref{sec:investigation_failure_modes}, we have shown that previous physics-based methods failed, which we investigated in detail. To proceed, let us first introduce an auxiliary flux parameter $\sigma(x) = -\alpha u_x(x)$ to reduce \eqref{eq:convection_diffusion_pde} to a system of first-order partial differential equations as follows 
\begin{subequations}
\begin{align}
    \mathcal{D}(x) &= \frac{d u(x)}{d x} + \frac{d \sigma(x)}{dx}, x \in (0,1),\\
    \mathcal{F}(x) &= \sigma(x) + \alpha \frac{d u(x)}{dx},x \in (0,1) 
\end{align}
\end{subequations}
where $\mathcal{D}$ is the residual form of our differential equation and $\mathcal{F}$ is the flux constraint along with Dirichlet boundary conditions $u(0) = 0.5$ and $u(1) = -0.5$. As mentioned in section \ref{sec:investigation_failure_modes}, we use a fully connected feed-forward neural network with four hidden layers and 20 neurons for this problem. Our network employs tangent hyperbolic non-linearity and has one input and two outputs corresponding to $u$ and $\sigma$. We adopt L-BFGS optimizer with its default parameters, built-in PyTorch framework \cite{paszke2019pytorch} and train our network for 5000 epochs. We generate $2048$ number of collocation points in the domain only once before training. To investigate the accuracy of our model, we present the results of our numerical experiment in Figure~\ref{fig:proposed_convection_diffusion}. 

Our model successfully learned the underlying solution as shown in Figure~\ref{fig:proposed_convection_diffusion}(a). From Figure~\ref{fig:proposed_convection_diffusion}(b), we observe that our proposed model adaptively focused on regions in which flux constraint was challenging to enforce. Moreover, from Figure~\ref{fig:proposed_convection_diffusion}(c), we observe that our proposed approach produces smooth loss landscapes with a visible minimum that our optimizer was able to find.

\subsection{Unsteady Heat Transfer in Composite Medium}
In this section, We study a typical heat transfer in a composite material where temperature and heat fluxes are matched across the interface \cite{baker1985heat}. Consider a time-dependent heat equation in a composite medium,
\begin{align}
   \frac{\partial T(x,t)}{\partial t} &= \frac{\partial }{\partial x}[a(x,t) \frac{\partial T(x,t)}{\partial x}] +  f(x,t), \quad (x,t) \in \Omega \times [0,\tau]
    \label{eq:heat_pde}
\end{align}
along with the Dirichlet boundary condition 
\begin{align}
    T(x,t) = g(x,t), \quad (x,t) \in \partial \Omega \times (0,\tau],
\end{align}
and initial condition 
\begin{align}
    T(x,0) = h(x),  \quad x \in \Omega
\end{align}
where $T$ is the temperature, $f$ is a heat source function, $a$ is thermal conductivity, $g$ and $h$ are source functions respectively. For non-smooth $T(x,t)$ and $a(x,t)$, we cannot directly apply \eqref{eq:heat_pde} due to the stringent smoothness requirement of our differential equation. For this reason, we relax the stringent smoothness requirement by introducing an auxiliary flux parameter $\sigma(x,t) = - a(x,t) \frac{\partial T(x,t)}{\partial x}$ to obtain a system of first-order partial differential equation that reads
\begin{subequations}
\begin{align}
\mathcal{D}(x,t) &= \frac{\partial T(x,t)}{\partial t} -  \frac{\partial \sigma(x,t) }{\partial x} +  f(x,t), \in \Omega \times [0,\tau],\\
\mathcal{F}(x,t) &= \sigma(x,t) + a(x,t) \frac{\partial T(x,t)}{\partial x}, \in \Omega \times [0,\tau],\\
\mathcal{B}(x,t) &= T(x,t) - g(x,t), \in \partial \Omega \times (0,\tau],\\
\mathcal{I}(x,t) &= T(x,0) - h(x), \in \Omega, t = 0,
\end{align}
\end{subequations}
where $\mathcal{D}$ is our differential equation, $\mathcal{F}$ is our flux constraint, $\mathcal{B}$ is our boundary condition constraint, and $\mathcal{I}$ is our initial condition constraint. We perform a numerical experiment in a composite medium of two non-overlapping sub-domains where $\Omega = \Omega_1 \cup \Omega_2$. We consider the thermal conductivity of the medium to vary as follows
\begin{equation}
a(x,t) = 
\begin{cases} 
      1, & (x,t) \in \Omega_1 \times [0,2] \\
      3 \pi, & (x,t) \in \Omega_2 \times [0,2] 
   \end{cases}
\end{equation}
where $\Omega_1 = \{x | -1 \le x < 0 \}$ and $\Omega_2 = \{x | 0 < x \le 1 \}$. To accurately evaluate our model, we consider an exact solution of the form
\begin{equation}
T(x,t) = 
\begin{cases} 
      \sin(3 \pi  x)  t, & x \in \Omega_1 \times [0,2] \\
      t  x, & x \in \Omega_2 \times [0,2].
   \end{cases}
   \label{eq:exact_heat_solution}
\end{equation}

The corresponding source functions $f(x,t)$, $g(x,t)$, and $h(x,t)$ can be calculated exactly using \eqref{eq:exact_heat_solution}. We use a fully connected neural network architecture consisting of four hidden layers with 20 neurons and tangent hyperbolic activation functions. We generate $N_{\Omega} = 2028$ collocation points from the interior part of the domain, $N_{\partial \Omega} = 2 \times 512$ on boundaries, and $N_{\mathcal{I}} = 512$ for approximating the initial conditions only once before training. We use L-BFGS  optimizer \cite{nocedal1980updating} with its default parameters and \emph{strong Wolfe} line search function that is built in PyTorch framework \cite{paszke2019pytorch}. We train our network for 10000 epochs.

\begin{figure}[!ht]
 \centering
    \subfloat[]{\includegraphics[scale=0.5]{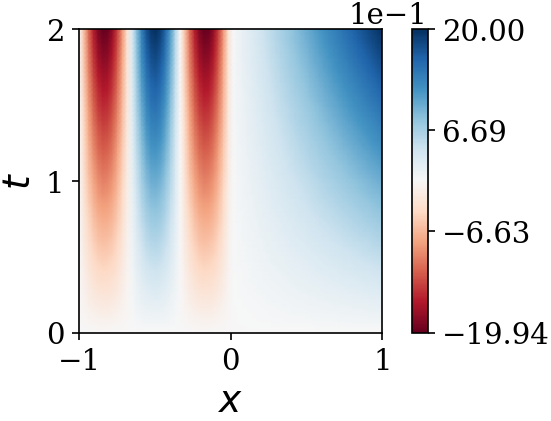}}\quad
    \subfloat[]{\includegraphics[scale=0.5]{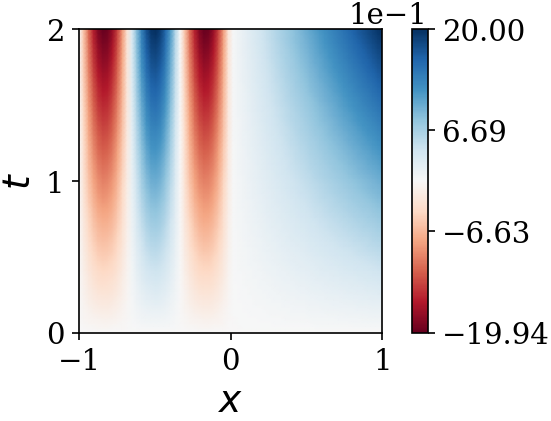}}\quad
    \subfloat[]{\includegraphics[scale=0.47]{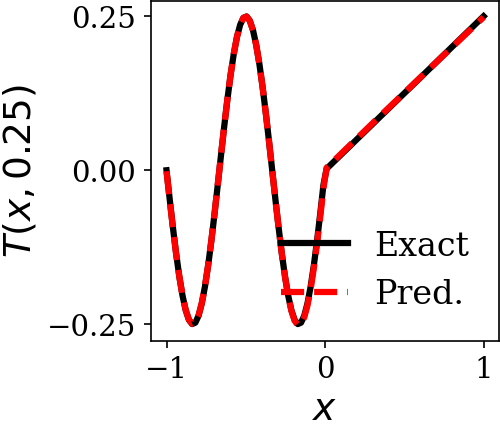}} \\
    \subfloat[]{\includegraphics[scale=0.5]{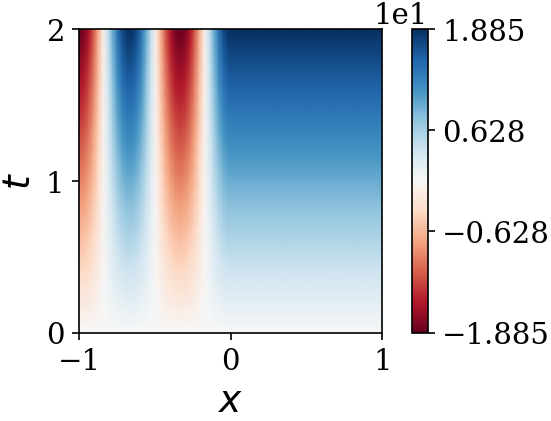}}\quad
    \subfloat[]{\includegraphics[scale=0.5]{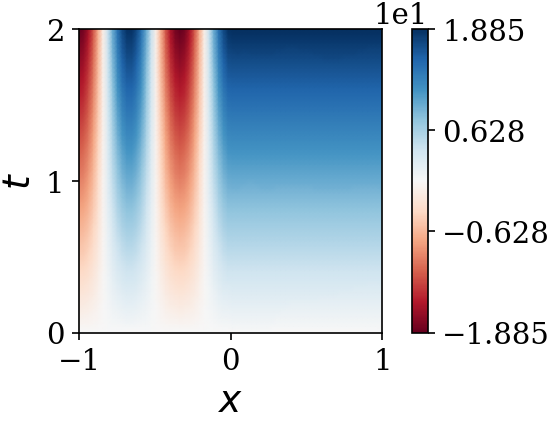}}\quad
    \subfloat[]{\includegraphics[scale=0.47]{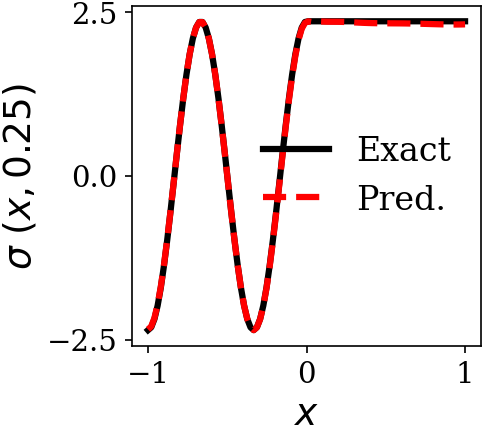}} \\
    
    \caption{Composite medium heat: Top row: $\mathcal{E}_r = 4.223 \times 10^{-3}$ (a) predicted solution, (b) exact solution (c) exact vs. predicted solution over the line. Bottom row: $\mathcal{E}_r = 1.536\times 10^{-3}$ (d) predicted flux, (e) exact flux (f) exact vs. predicted flux over line }
    \label{fig:proposed_unsteady_heat_composite}
\end{figure}

The result of the experiment is summarized in Figure~\ref{fig:proposed_unsteady_heat_composite}. We observe that our neural network model has successfully learned the underlying solution as shown in Figures~\ref{fig:proposed_unsteady_heat_composite}(a)-(b)-(c).
 Similarly, our model successfully learned to predict the flux distribution, as can be seen in Figures~\ref{fig:proposed_unsteady_heat_composite}(d)-(e)-(f).

\subsection{Convection Equation}
In this section, we study the transport of a physical quantity dissolved or suspended in an inviscid fluid with a constant velocity. Numerical solutions of inviscid flows often exhibit a viscous or diffusive behavior owing to numerical dispersion. Let us consider a convection equation of the form

\begin{align}
        \frac{\partial \xi}{\partial t} + u \frac{\partial \xi}{\partial x} &= 0, ~\forall (x,t) \in \Omega \times [0,1],
        \label{eq:convectionPDE}
\end{align}
satisfying the following boundary condition 
\begin{align}
    \xi(0,t) &= \xi(2 \pi ,t) ~\forall t \in [0,1]
\end{align}
and initial condition
\begin{align}
        \xi(x,0) &= h(x), ~ \forall x \in \partial \Omega.
\end{align}
where $\xi$ is any physical quantity to be convected with the velocity $u$,  $\Omega = \{x~|~ 0 < x < 2\pi \}$ and $\partial \Omega$ is its boundary. Eq.\eqref{eq:convectionPDE} is inviscid, so it lacks viscosity or diffusivity. 
For this problem, we consider $u=40$ and  $h(x) = \sin(x)$. The analytical solution for the above problem is given as follows \cite{krishnapriyan2021characterizing}
\begin{equation}
    \xi(x,t) = F^{-1}(F(h(x) )e^{-i u k t}),
\end{equation}
where $F$ is the Fourier transform, $k$ is the frequency in the Fourier domain and $i = \sqrt{-1}$. Since the PDE is already first-order, there is no need to introduce any auxiliary parameters. Following is the residual form of the PDE used to formulate our objective function,
\begin{subequations}
\begin{align}
    &\mathcal{D}(x,t) = \frac{\partial \xi(x,t)}{\partial t}
    + u \frac{\partial \xi(x,t)}{\partial x}, \\
    &\mathcal{B}(t)= \xi(0,t) - \xi(2\pi,t),\\
    &\mathcal{I}(x) = \xi(x,0) - \sin(x),
\end{align}
\label{eq:residual_form_convection_equation}
\end{subequations}
where $\mathcal{D}$ is the residual form of our differential equation, $\mathcal{B}$ and $\mathcal{I}$ are our boundary condition and initial condition constraints. As we discussed in section \ref{sec:PINNs}, \cite{krishnapriyan2021characterizing} proposed curriculum learning for the solution of \eqref{eq:convectionPDE}. For this particular problem, the complexity of learning the solution $\xi(x,t)$ increases with increasing $u$. However, we re-emphasize that for a general PDE, it is known a priori what factors control the complexity of the solution. 

\begin{figure}[!ht]
 \centering
    \subfloat[]{\includegraphics[scale=0.46]{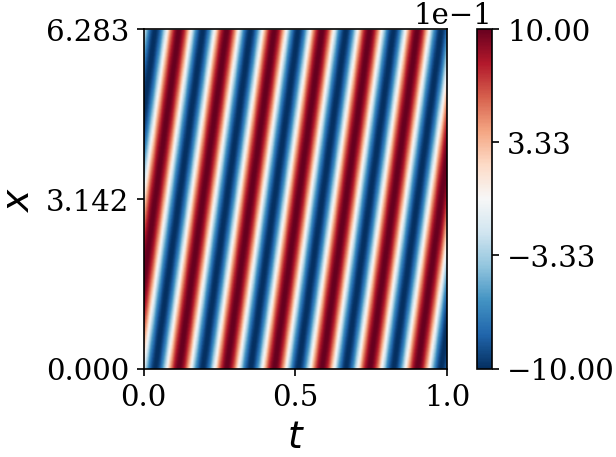}}\quad
    \subfloat[]{\includegraphics[scale=0.46]{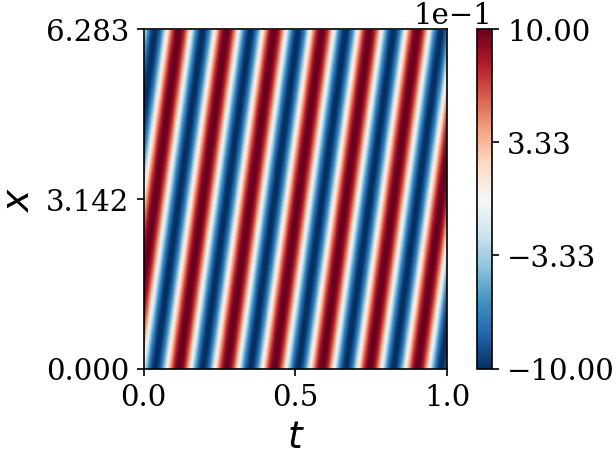}}\quad
    \subfloat[]{\includegraphics[scale=0.46]{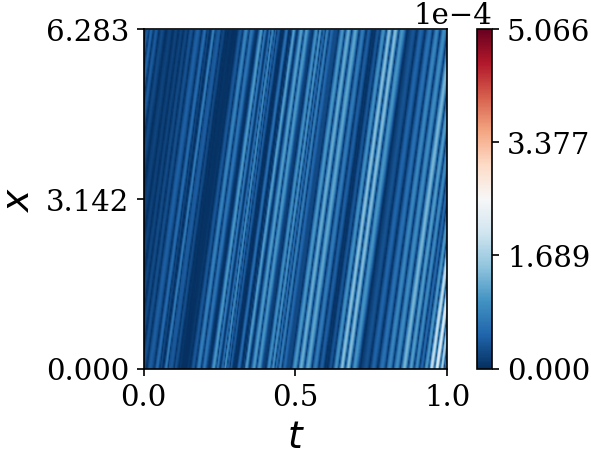}}
    \caption{Convection equation: (a) exact solution, (b) predicted solution, (c) absolute point-wise error}
    \label{fig:proposed_convection_equation}
\end{figure}

For this problem, we use a fully connected neural network architecture consisting of four hidden layers with 50 neurons and tangent hyperbolic activation functions. We generate $N_{\Omega} = 2048$ collocation points from the interior part of the domain, $N_{\partial \Omega} = 512$ from each boundary, and $N_{\mathcal{I}} = 512$ for approximating the initial conditions only once before training. We use L-BFGS  optimizer \cite{nocedal1980updating} with its default parameters and \emph{strong Wolfe} line search function that is built in PyTorch framework \cite{paszke2019pytorch}. We train our network for 10000 epochs. We present the prediction of our neural network in Figure~\ref{fig:proposed_convection_equation}. We observe that our neural network model has successfully learned the underlying solution as shown in Figures~\ref{fig:proposed_convection_equation}(b)-(c). 

\begin{table}[!ht]
\centering
\caption{Convection equation: summary of the $\mathcal{E}_r$ and MAE errors for training a fixed neural network architecture with different methods along}
\label{tb:Convection}
\vspace{2pt}
\resizebox{0.65\textwidth}{!}{%
\begin{tabular}{@{}lccccc@{}}
\toprule
\multicolumn{1}{c}{Models} &
  \multicolumn{1}{c}{$\mathcal{E}_r(\xi,\hat{\xi})$} &
  \multicolumn{1}{c}{MAE} &
  \\ \midrule
Curriculum Learning \cite{krishnapriyan2021characterizing}  & $5.33 \times 10^{-2}$& $2.69 \times 10^{-2}$ \\
Proposed method & $\boldsymbol{5.787 \times 10^{-5}}$ & $\boldsymbol{4.656 \times 10^{-5}}$ \\
\end{tabular}}
\end{table}

We also present a summary of the error norms from our approach and state-of-the-art results given in \cite{krishnapriyan2021characterizing} in Table~\ref{tb:Convection}. We observe that our method achieves a relative $\mathcal{E}_r = 5.787 \times 10^{-5}$, which is three orders of magnitude better than $5.33 \times 10^{-2}$ obtained by the method presented in \citet{krishnapriyan2021characterizing}.

\subsection{Stokes Equation}
The Stokes problem applies to many branches of physics and engineering. Stokes operators are fundamental to more complicated models of physical phenomena such as Navier-Stokes equations. In this numerical example, we consider a two-dimensional Stokes equation. Following the approach in section \ref{sec:Proposed_Method}, we transform our partial differential equation into a system of first-order PDE as follows
\begin{subequations}
\begin{align}
    \mathcal{D}_x &=\frac{\partial p}{\partial x} + \frac{1}{Re}\frac{\partial \omega}{\partial y} - f_x,\\
    \mathcal{D}_y &=\frac{\partial p}{\partial y} - \frac{1}{Re}\frac{\partial \omega}{\partial x} -f_y,\\
     \mathcal{F} &= \frac{\partial v}{\partial x} -\frac{\partial u}{\partial y} - \omega,\\
     \mathcal{C} &=\frac{\partial u}{\partial x} + \frac{\partial v}{\partial y}
\end{align}
\end{subequations}
where $\omega$ is the vorticity, $Re$ is the Reynolds number, $(u,v)$ is the velocity vector, $p$ is the pressure, and $(f_x, f_y) $ is the body force vector. $\mathcal{D} = (\mathcal{D}_x,\mathcal{D}_y)$ is the residual form of our differential equation, $\mathcal{F}$ is our vorticity constraint, and $\mathcal{C}$ is our divergence constraint. We consider a model problem $Re = 1$ studied by Oden and Jacquotte\cite{oden1984stability} that corresponds to the divergence-free velocity field 
\begin{align}
    u(x,y) = x^2(1-x)(2y-6y^2 + 4y^3), \quad
    v(x,y) = y^2(1-y)^2(-2x + 6x^2 -4x^3),
\end{align}
the pressure field
\begin{align}
    p(x,y) = x^2 - y^2,
\end{align}
the vorticity field
\begin{align}
    \omega(x,y) =-x^2(1-x)^2(2- 12y + 12y^2)+y^2(1-y)^2(-2 +12x - 12x^2),
\end{align}
and the following boundary conditions
\begin{subequations}
\begin{align}
    u(x,y) &= 0, v(x,y) = 0 ~\in \{(x,y)|~x = 0,~ 0 \le y \le 1\}, \\
    u(x,y) &= 0, v(x,y) = 0  ~\in \{(x,y)| x = 1,~ 0 \le y \le 1\},\\
    u(x,y) &= 0, v(x,y) = 0  ~\in \{(x,y)| 0 \le x \le 1,~ y = 0\},\\
    u(x,y) &= 0, p(x,y) = x^2-1 ~\in \{(x,y)| 0 \le x \le 1, ~y = 1\},
\end{align}
\end{subequations}
We use a fully connected neural network architecture consisting of four hidden layers with 50 neurons and tangent hyperbolic activation functions for this problem. We generate $N_{\Omega} = 2048$ collocation points from the interior part of the domain, $N_{\partial \Omega} = 512$ from each boundary face only once before training. We choose L-BFGS optimizer \cite{nocedal1980updating} with its default parameters and \emph{strong Wolfe} line search function that is built in PyTorch framework \cite{paszke2019pytorch}. Finally, we train our network for 10000 epochs.
\begin{figure}[!h]
\centering
    \subfloat[]{\includegraphics[scale=0.48]{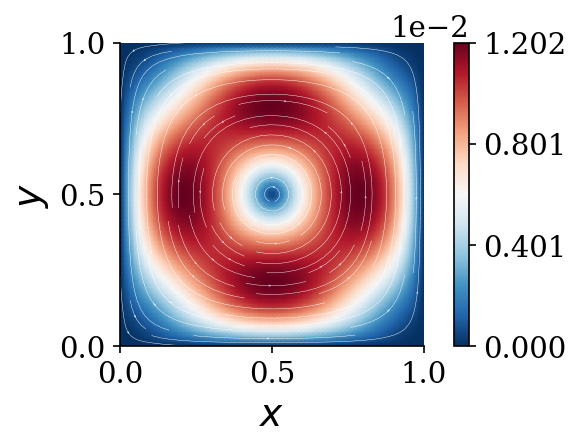}}
    \subfloat[]{\includegraphics[scale=0.48]{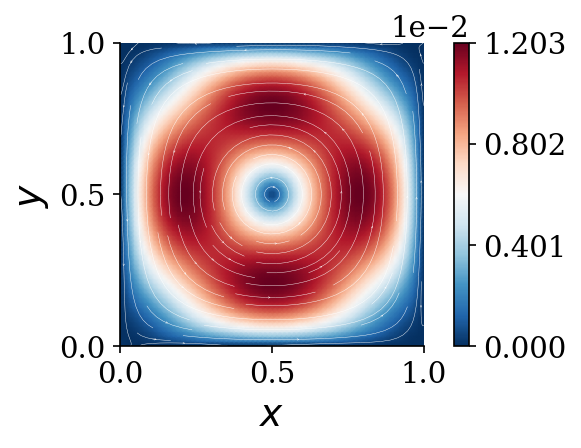}}
    \subfloat[]{\includegraphics[scale=0.48]{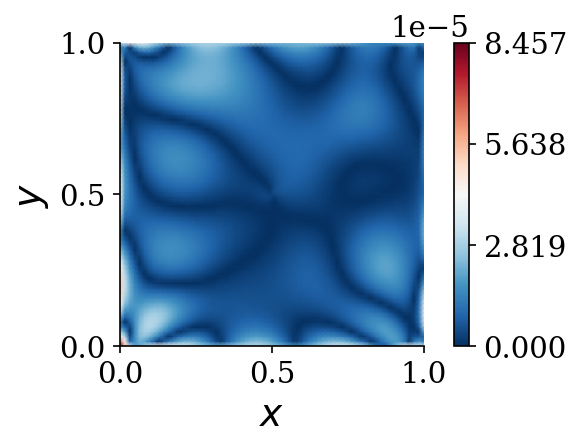}} \\
    \subfloat[]{\includegraphics[scale=0.48]{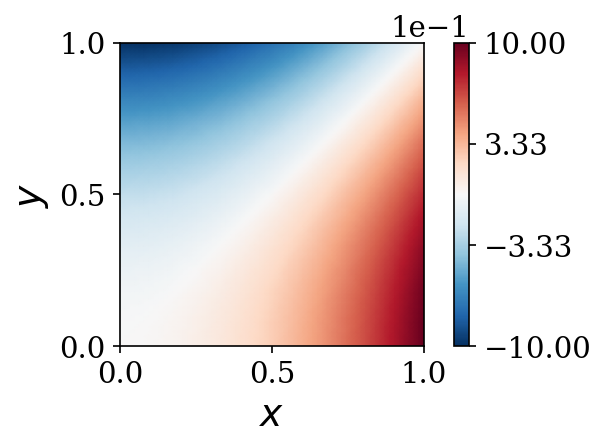}}
    \subfloat[]{\includegraphics[scale=0.48]{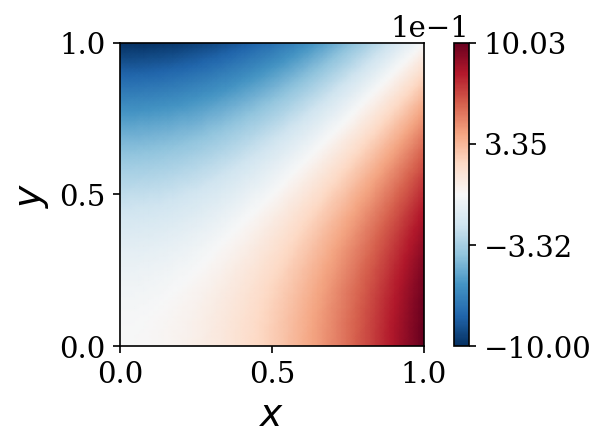}}
    \subfloat[]{\includegraphics[scale=0.48]{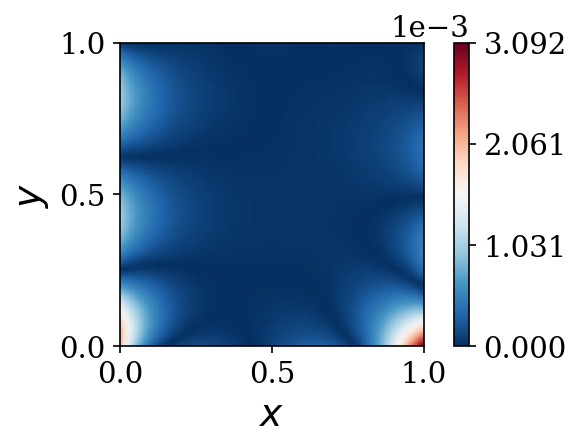}} \\
    \subfloat[]{\includegraphics[scale=0.48]{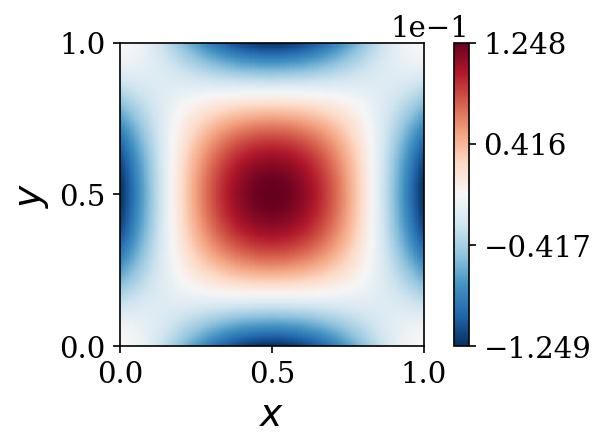}}
    \subfloat[]{\includegraphics[scale=0.48]{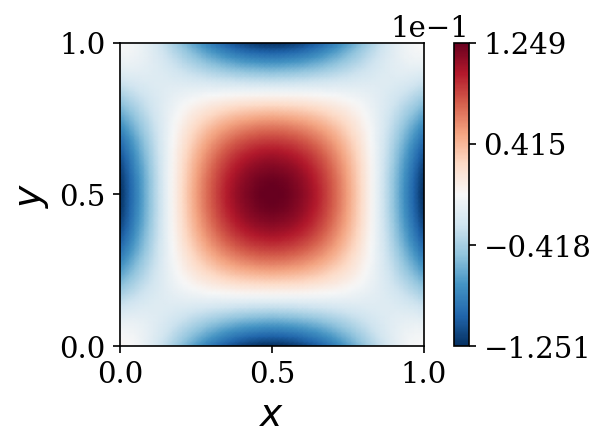}}
    \subfloat[]{\includegraphics[scale=0.48]{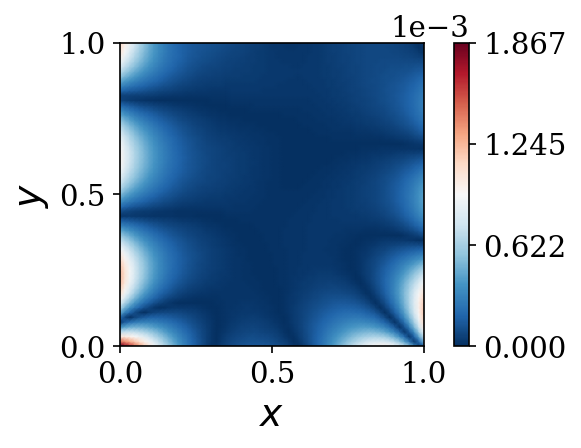}} 
    
    \caption{Stokes equation: Top row: velocity with $\mathcal{E}_r(\boldsymbol{u},\hat{\boldsymbol{u}}) = 1.050 \times 10^{-3}$. (a) exact velocity field, (b) predicted velocity field, and (c) absolute point-wise error between predicted and exact speed. Middle row: pressure with $\mathcal{E}_r(p,\hat{p}) = 9.952 \times 10^{-4}$. (d) exact pressure distribution, (e) predicted pressure distribution and (f) absolute point-wise error between the exact and the predicted pressure. Bottom row: vorticity with $\mathcal{E}_r(\omega,\hat{\omega}) = 4.052 \times 10^{-3}$. (g) exact vorticity field, (h) predicted vorticity field, (i) absolute point-wise error between predicted and exact vorticity}
    \label{fig:proposed_stokes_vorticity_field}
\end{figure}

 Results from our numerical experiments are summarized in Figure~\ref{fig:proposed_stokes_vorticity_field}. Our model successfully learned the underlying solution accurately, as shown in Figures~\ref{fig:proposed_stokes_vorticity_field}(b)-(c). Similarly, we observe excellent agreement of the predicted pressure with the exact pressure field as depicted in Figures~\ref{fig:proposed_stokes_vorticity_field}(e)-(f). In addition, the predicted vorticity field from our model perfectly matches the exact vorticity field as illustrated in Figures~\ref{fig:proposed_stokes_vorticity_field}(g)-(h)-(i)
Next, we conduct a benchmark numerical experiment using the incompressible Navier-Stokes Equation.
\subsection{Incompressible Navier-Stokes Equation}
In this section, we study a classical benchmark problem in computational fluid dynamics, the steady-state flow in a two-dimensional lid-driven cavity. The system is governed by the incompressible Navier-Stokes equations, which can be written in a non-dimensional form as
\begin{subequations}
\begin{align}
    \frac{\partial u}{\partial x} + \frac{\partial v}{\partial y} &=0, \quad (x,y) \in \Omega \\
    u \frac{\partial u}{\partial x} + v \frac{\partial u}{\partial y} + \frac{\partial p}{\partial x} - \frac{1}{Re}(\frac{\partial^2 u}{\partial x^2} + \frac{\partial^2 u}{\partial y^2}) &=f_x, \quad (x,y) \in \Omega\\
    u \frac{\partial v}{\partial x} + v \frac{\partial v}{\partial y} +  \frac{\partial p}{\partial y} - \frac{1}{Re}(\frac{\partial^2 v}{\partial x^2} + \frac{\partial^2 v}{\partial y^2})  &=f_y,\quad (x,y) \in \Omega
\end{align}
\label{eq:Lid_driven_cavity}
\end{subequations}
where $(u,v)$ is the velocity vector, $(f_x,f_y)$ is the body force vector, $Re = 100$ is the Reynolds number and $p$ is the pressure. We aim to solve the above equation on $\Omega = \{(x,y)~|~ 0 \le x \le 1, 0 \le y \le 1 \}$ with its boundary $\partial \Omega$. Our top wall moves at a velocity of $u(x,1) = 1$ in the x direction. The other three walls are applied as no-slip conditions. Because we have a closed system at a steady state with no inlets or outlets in which the pressure level is defined, we provide a reference level for the pressure $p(0,0) = 0$. The main challenge here is to solve the above problem using the velocity-pressure formulation.

\citet{wang2021understanding} reported that their method failed to learn the underlying solution using pressure-velocity formulation. \citet{cpinns2020} used 12 networks and $2500 \times 16$ number of collocation points with more than a thousand boundary points to learn the underlying solution. In this work, we use only $2500$ collocation points, and only one of the networks in \cite{cpinns2020} to learn the underlying solution accurately. The proposed method presented in section \ref{sec:Proposed_Method} can be generalized to transform \eqref{eq:Lid_driven_cavity} into a system of first-order velocity-pressure-vorticity formulation,

\begin{subequations}
\begin{align}
    \mathcal{C}(x,y) &= \frac{\partial u}{\partial x} + \frac{\partial v}{\partial y}, \quad (x,y) \in \Omega\\
    \mathcal{D}_x(x,y) &= u \frac{\partial u}{\partial x} + v \frac{\partial u}{\partial y} + \frac{\partial p}{\partial x} + \frac{1}{Re}\frac{\partial \omega}{\partial y} - f_x,\quad (x,y) \in \Omega\\
    \mathcal{D}_y(x,y) &=u \frac{\partial v}{\partial x} + v \frac{\partial v}{\partial y} +  \frac{\partial p}{\partial y} - \frac{1}{Re}\frac{\partial \omega}{\partial x} - f_y,\quad (x,y) \in \Omega\\
     \mathcal{F}(x,y) &=\frac{\partial v}{\partial x} -\frac{\partial u}{\partial y}- \omega, \quad (x,y) \in \Omega
\end{align}
\end{subequations}

\begin{figure}[!ht]
 \centering
    \subfloat[]{\includegraphics[scale=0.50]{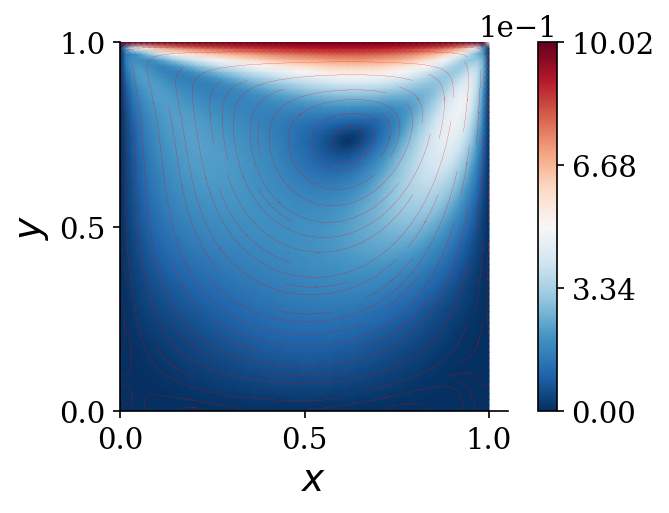}}
    \subfloat[]{\includegraphics[scale=0.52]{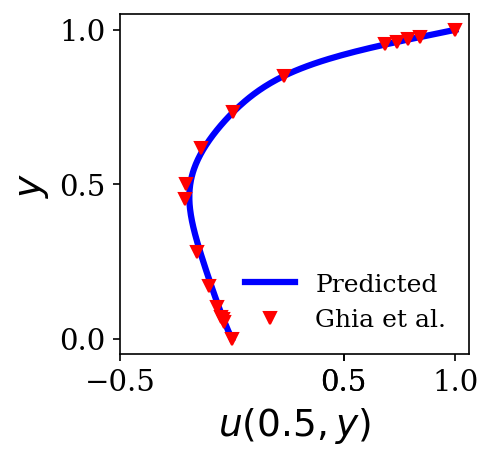}}
    \subfloat[]{\includegraphics[scale=0.52]{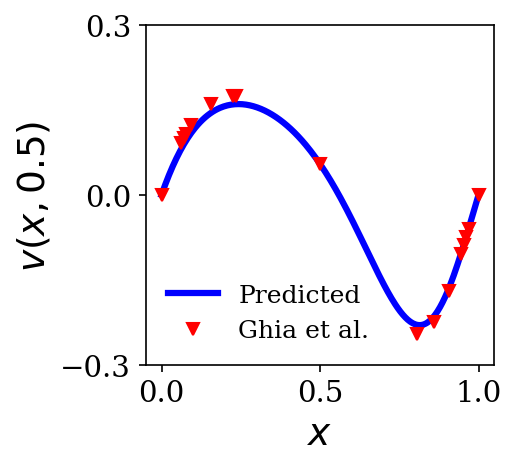}} 
    \caption{Navier-Stokes equation: (a) predicted solution, (b) predicted horizontal velocity over line compared with the benchmark result by, (c) predicted vertical velocity over line compared with the benchmark result by }
    \label{fig:proposed_navier_stokes}
\end{figure}

Where $\omega$ is the vorticity field, $\mathcal{D} = (\mathcal{D}_x,\mathcal{D}_y)$ is the residual form of our differential equation, $\mathcal{F}$ is our vorticity constraint, and $\mathcal{C}$ is our divergence constraint. We use a fully connected neural network architecture of six hidden layers with 20 neurons and tangent hyperbolic activation functions. We generate $N_{\Omega} = 2500$ collocation points uniformly from the interior part of the domain, $N_{\partial \Omega} = 4\times128$ number of points on the boundaries only once before training. In addition, we constrain our pressure at the corner $(0,0)$. We choose L-BFGS optimizer \cite{nocedal1980updating} with its default parameters and \emph{strong Wolfe} line search function that is built in PyTorch framework \cite{paszke2019pytorch}. We train our network for 20000 epochs. We present the prediction of our neural network along with benchmark results \cite{ghia_1982} in Figure~\ref{fig:proposed_navier_stokes}. 
From Figures~\ref{fig:proposed_navier_stokes}(b)-(c), we observe that the predictions of our neural network model are in excellent agreement with the benchmark results reported in \cite{ghia_1982}.

\section{Conclusion}\label{sec:Conclusion}

In this paper, we explored the limitations of using physics-informed neural networks (PINNs) for solving partial differential equations (PDEs). We highlighted that the absence of prior knowledge of the solution or validation data makes it difficult to adjust the hyperparameters, thereby rendering using PINNs impracticable for solving forward problems. We found that existing methods struggle with high-order PDEs and produce complex loss landscapes that are difficult to optimize. We also showed that backpropagated gradients are contaminated by high-order differential operators, resulting in unpredictable training. Furthermore, the strong form of governing PDEs may hinder learning PDEs with non-smooth solutions. To overcome these challenges, we proposed a novel method that reduces the order of a given PDE via auxiliary variables and formulated an unconstrained optimization problem using Lagrange multiplier method. We learned our primary and auxiliary variables using a single neural network model to promote learning shared hidden features and to reduce the risk of overfitting. We applied our method to solve several benchmark problems, including a convection-dominated convection-diffusion, convection equation, and incompressible Navier-Stokes equation. We demonstrated marked improvements over existing neural network-based methods. In future research, our method will be applied to tackle three-dimensional fluid flows with high Reynolds numbers. We plan to further investigate the impact of physics on contaminating backpropagated gradients to develop improved initialization schemes for enhancing the trainability of neural networks on physics.

\appendix
\section{Sensitivity Analysis of Backpropagated Gradients in Presence of Physics}\label{sec:appendix_sensitivity_gradients}
In section \ref{sec:effect_of_Diff_Opt}, we discussed how differential operators amplify the embedded noise in the output of a neural network model trained with L-BFGS optimizer. In this section, we investigate the contamination of backpropagated gradients during training a PINN model for the solution of \eqref{eq:convection_diffusion_pde} with Adam optimizer. 
\begin{figure}[!h]
\centering
    \subfloat[]{\includegraphics[scale=0.40]{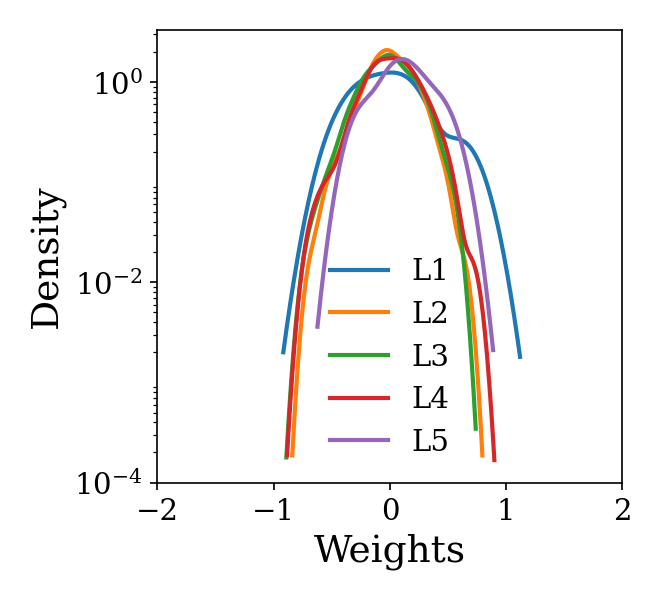}}\qquad
    \subfloat[]{\includegraphics[scale=0.40]{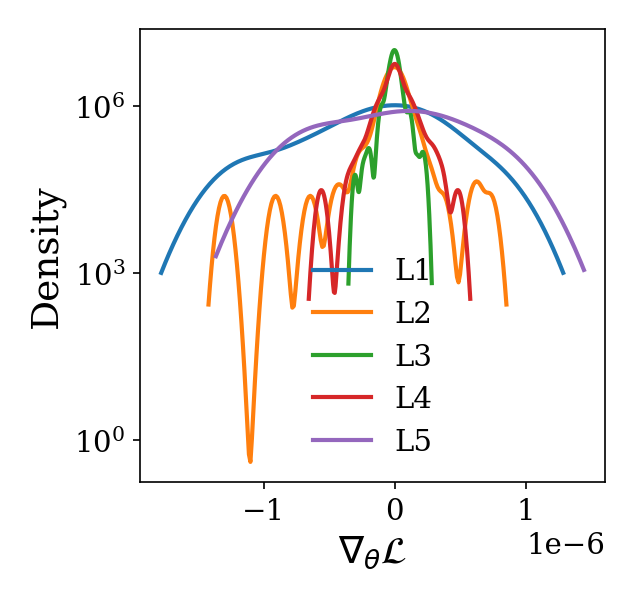}}\qquad
    \subfloat[]{\includegraphics[scale=0.40]{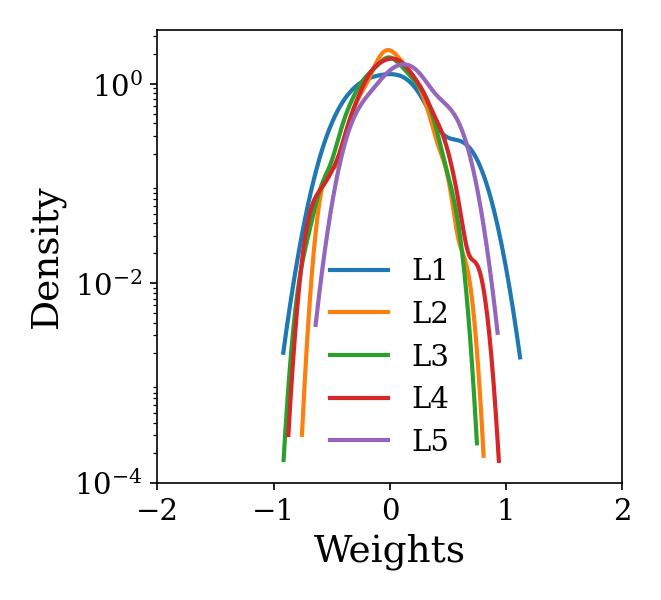}}\qquad
    \subfloat[]{\includegraphics[scale=0.40]{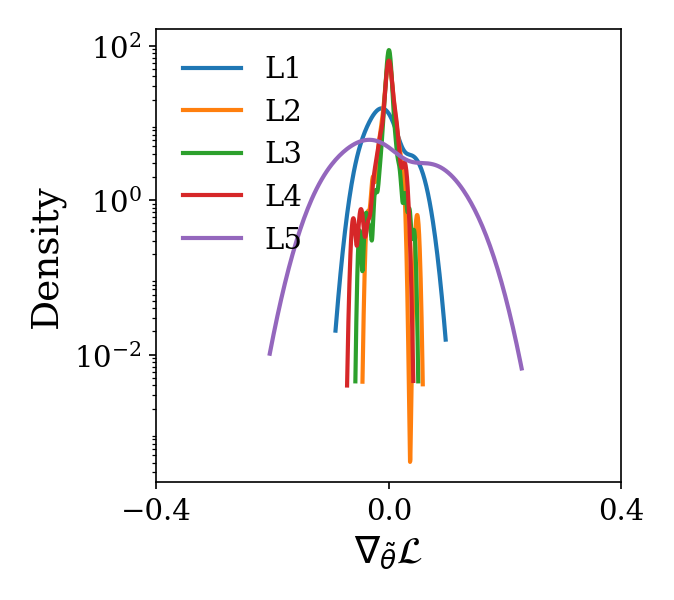}}\qquad
    \subfloat[]{\includegraphics[scale=0.40]{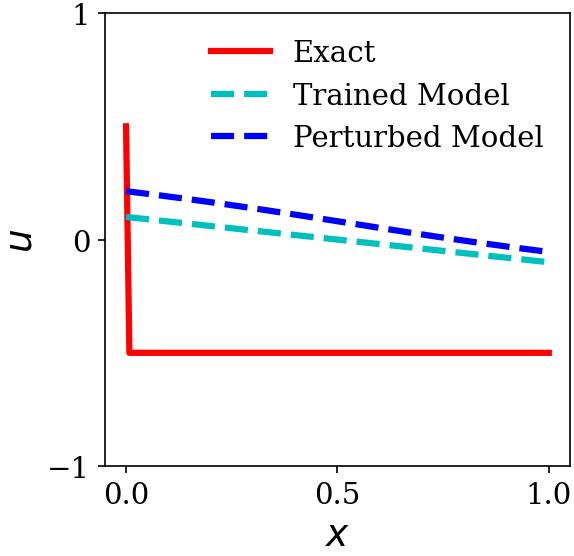}}
    \caption{effect of a differential operator in training our PINN model with Adam optimizer:(a) distribution of parameters of the network before perturbation, (b) distribution of gradients of the parameters of the network before perturbation, (c) distribution of parameters of the network at the perturbed state, (d) distribution of gradients of the parameters of the network at the perturbed state, (e) prediction before and after perturbation}
    \label{fig:noise_amplification_pinn_model_adam}
\end{figure}
From Figures~\ref{fig:noise_amplification_pinn_model_adam}(a)-(c), we observe that perturbations are acceptable since the distribution of the parameters before and after perturbations are similar. However, backpropagated gradients have increased by almost several orders of magnitude, as can be seen from Figures~\ref{fig:noise_amplification_pinn_model_adam}(b)-(d). We also observe the impact of perturbations in predictions of our PINN model in Figures~\ref{fig:noise_amplification_pinn_model_adam}(e).
Similarly, we present the results of our numerical
experiment by training a PECANN model for the solution of \eqref{eq:convection_diffusion_pde} with Adam optimizer.
\begin{figure}[!h]
\centering
    \subfloat[]{\includegraphics[scale=0.40]{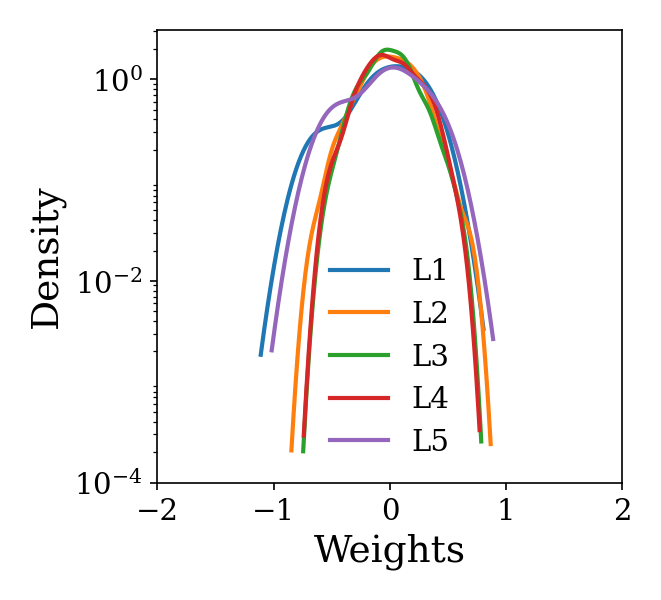}}\qquad
    \subfloat[]{\includegraphics[scale=0.40]{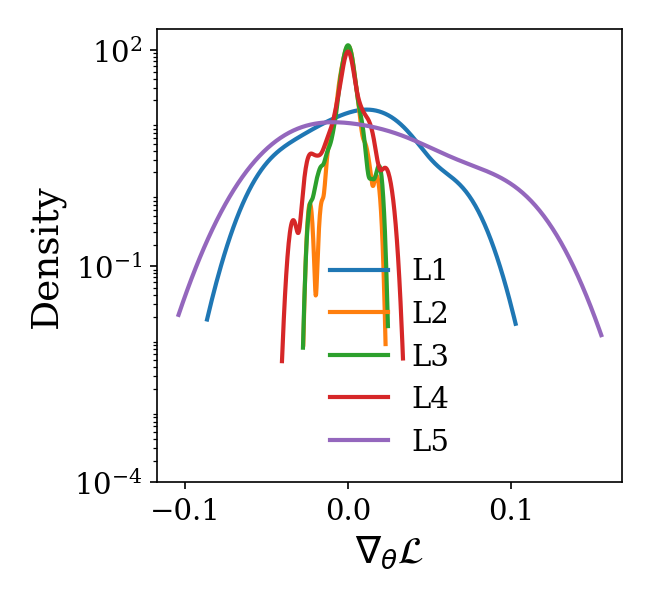}}\qquad
    \subfloat[]{\includegraphics[scale=0.40]{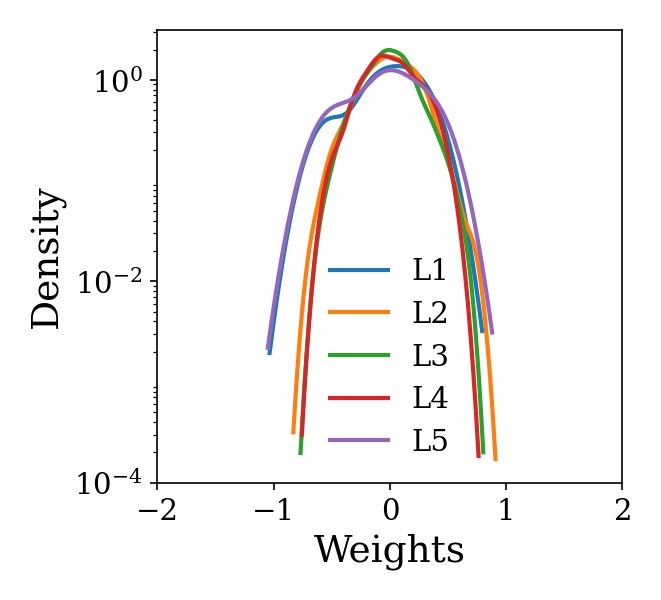}}\qquad
    \subfloat[]{\includegraphics[scale=0.40]{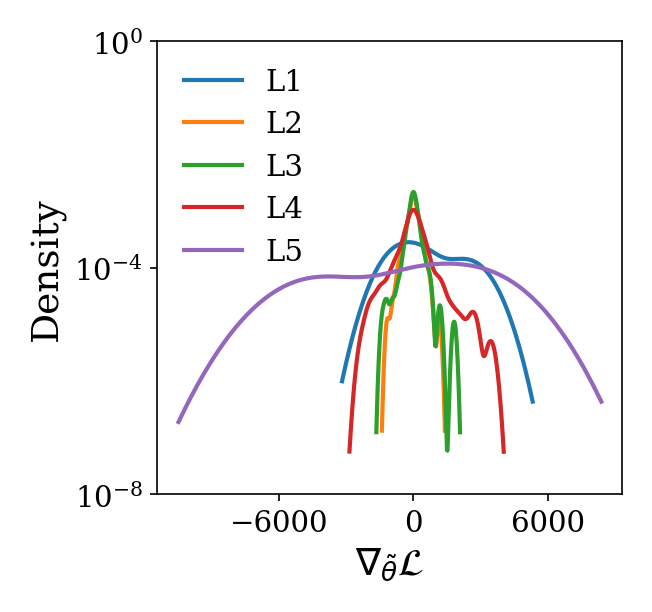}}\qquad
    \subfloat[]{\includegraphics[scale=0.40]{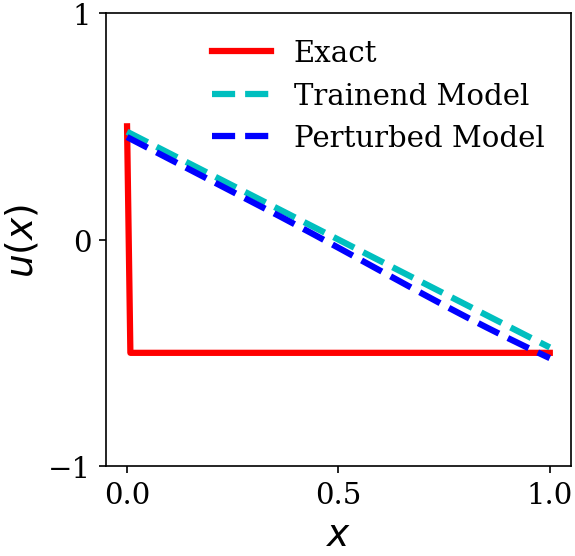}}\qquad
    \caption{effect of a differential operator in training our PECANN model with Adam optimizer:(a) distribution of parameters of the network before perturbation, (b) distribution of gradients of the parameters of the network before perturbation, (c) distribution of parameters of the network at the perturbed state, (d) distribution of gradients of the parameters of the network at the perturbed state, (e) prediction before and after perturbation}
    \label{fig:noise_amplification_pecann_model_adam}
\end{figure}
From Figures~\ref{fig:noise_amplification_pecann_model_adam}(a)-(c), we observe that perturbations are acceptable since the distribution of the parameters before and after perturbations are similar. However, backpropagated gradients have increased by several orders of magnitude, as can be seen from Figures~\ref{fig:noise_amplification_pinn_model_adam}(b)-(d). 
\subsection{Helmholtz Equation}
\label{sec:Helmholtz}
\begin{figure}[!ht]
 \centering
    \subfloat[]{\includegraphics[scale=0.5]{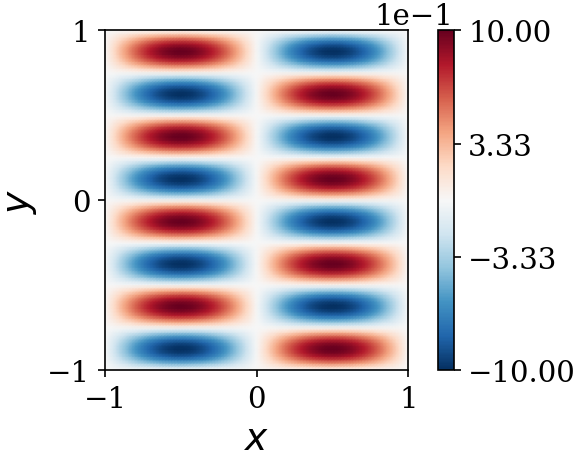}}
    \subfloat[]{\includegraphics[scale=0.5]{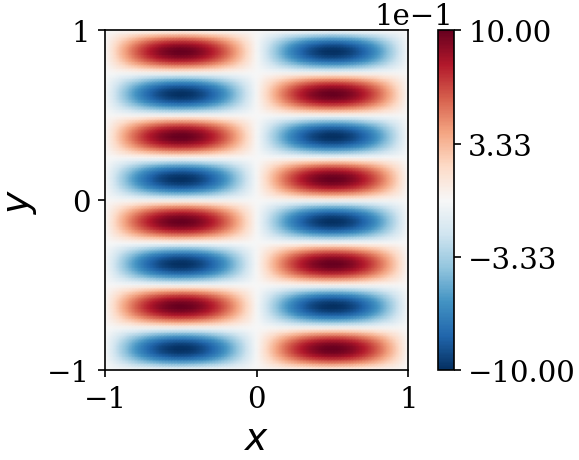}}
    \subfloat[]{\includegraphics[scale=0.5]{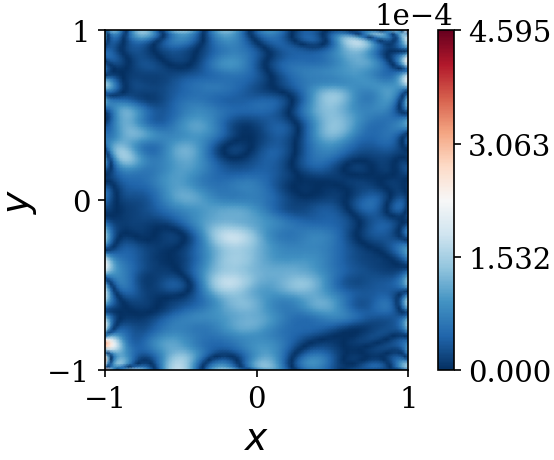}}
\caption{Helmholtz equation: (a) exact solution ,(b) predicted solution, (c) absolute point-wise error}
\label{fig:Helmholtz}
\end{figure}

In this section, we aim to demonstrate our proposed unconstrained formulation as presented in section \ref{sec:unconstrained_formuluation} for the solution of Helmholtz equation, which appears in various applications in physics and science \cite{plessix2007helmholtz,bayliss1985numerical,li2013three,greengard1998accelerating}. In two dimensions, our partial differential equation reads
\begin{align}
        \Delta u(x,y) + k^2 u(x,y) &= q(x,y), ~\forall (x,y) \in \Omega,
        \label{eq:HelmholtzPDE}
\end{align}
along with Dirichlet boundary conditions 
\begin{equation}
    u(x,y) = 0, ~ \forall (x,y) \in \partial \Omega,
    \label{eq:HelmholtzBC}
\end{equation}
where $k=1$,  $\Omega = \{ (x,y) ~ | ~ -1 \le x \le 1, -1 \le y \le 1 \}$ and $\partial \Omega$ is its boundary. It can be verified that \eqref{eq:HelmholtzPDE} and its boundary conditions as given in \eqref{eq:HelmholtzBC} can be solved with the following function
\begin{equation}
u(x,y) = \sin( \pi x)\sin(4 \pi y), ~\quad \forall (x,y) \in \Omega.
 \label{eq:HelmholtzExactSolution}
\end{equation}
This problem has been studied in \cite{wang2021understanding,mcclenny2020self}. For this problem, we use a fully connected neural network architecture as in \cite{wang2021understanding}, which consists of three hidden layers with 30 neurons per layer and the tangent hyperbolic activation function. We note that \cite{wang2021understanding} is generating their data at every epoch, which amounts to $N_{\Omega} = 5.12 \times 10^{6}$ and $N_{\partial \Omega} = 20.48 \times 10^{6}$. Similarly \cite{mcclenny2020self} generates $N_{\Omega} = 100 \times 10^3$ number of collocation points and $N_{\partial \Omega} = 400$. On the other hand, we use the Sobol sequence to uniformly generate $N_{\Omega} = 500$ residual points from the interior part of the domain and $N_{\partial \Omega} = 256$ from the boundaries only once before training. We note that our collocation points amount to only $0.5\%$ of the data generated in \cite{mcclenny2020self} and to $0.01\%$ of the data generated in \cite{wang2021understanding}. Our optimizer is L-BFGS  \cite{nocedal1980updating} with its default parameters and \emph{strong wolfe} line search function that is built in PyTorch framework \cite{paszke2019pytorch}. We train our network for 5000 epochs. The results of our numerical experiment are presented in Figure~\ref{fig:Helmholtz}, which shows that the prediction obtained from our model is indistinguishable from the exact solution. 

\begin{table}[!ht]
\centering
\caption{Helmholtz equation: the average and the standard deviations of the error norms over ten independent trials with the number of collocation points $N_{\Omega}$ and the number of boundary points $N_{\partial \Omega}$}
\label{tb:Helmholtz}
\vspace{2pt}
\resizebox{\textwidth}{!}{%
\begin{tabular}{@{}lccccc@{}}
\toprule
\multicolumn{1}{c}{Models} &
  \multicolumn{1}{c}{$\mathcal{E}_r(u,\hat{u})$}&
  \multicolumn{1}{c}{$\mathcal{E}_\infty(u,\hat{u})$} &
  \multicolumn{1}{c}{$N_{\Omega}$} &
  \multicolumn{1}{c}{$N_{\partial \Omega}$}
  \\ \midrule
Ref. \cite{wang2021understanding}  & $4.31 \times 10^{-2} \pm 1.68 \times 10^{-2}$& - & $128 \times 40000$ & $4 \times 128 \times 40000 $ \\
Proposed Method  & $\boldsymbol{2.77 \times 10^{-4} \pm 9.15 \times 10^{-5}}$ & $\boldsymbol{9.04 \times 10^{-4} \pm 3.77 \times 10^{-4}}$ & $500$ & $4 \times 64$&
\end{tabular}}
\end{table}

Furthermore, we report the summary of the error norms obtained from our approach with state-of-the-art results presented in \cite{wang2021understanding} averaged over ten independent trials with random Xavier initialization scheme\cite{glorot2010understanding} in Table~\ref{tb:Helmholtz}, which shows that our method outperforms the method presented in \citet{wang2021understanding} by two orders of magnitude in error levels with just $0.01\%$ of their generated collocation points.
\section*{Acknowledgments}
The author would like to thank Dr. Inanc Senocak for his valuable comments.

\section*{Funding}
This material is based upon work supported by the National Science Foundation under Grant No. 1953204 and in part by the University of Pittsburgh Center for Research Computing through the resources provided

\clearpage

\end{document}